\documentclass[sigconf]{acmart}
\usepackage{color}
\usepackage{newfloat}
\usepackage{listings}
\usepackage{algorithm}
\usepackage{algorithmic}
\usepackage{amsthm}
\usepackage{dsfont}
\usepackage{multirow}
\usepackage{amsmath}
\usepackage{amsfonts}
\usepackage{bm}
\usepackage{booktabs}
\usepackage{enumitem}

\AtBeginDocument{%
  }

\copyrightyear{2026}
\acmYear{2026}
\setcopyright{cc}
\setcctype{by}
\acmConference[WWW '26]{Proceedings of the ACM Web Conference 2026}{April 13--17, 2026}{Dubai, United Arab Emirates}
\acmBooktitle{Proceedings of the ACM Web Conference 2026 (WWW '26), April 13--17, 2026, Dubai, United Arab Emirates}
\acmPrice{}
\acmDOI{10.1145/3774904.3792252}
\acmISBN{979-8-4007-2307-0/2026/04}

\begin{document}

\title{Conditional Diffusion Guided Knowledge Transfer for Multi-Domain Knowledge Graph Completion}

\author{Jiawei Sheng}
\affiliation{%
    \institution{Institute of Information Engineering, \\ Chinese Academy of Sciences} 
    \state{Beijing}
    \country{China}
}
\email{shengjiawei@iie.ac.cn}
\orcid{0000-0002-4865-982X}

\author{Taoyu Su}
\affiliation{%
    \institution{Institute of Information Engineering, \\ Chinese Academy of Sciences} 
    \state{Beijing}
    \country{China}
}
\authornote{Corresponding author.}
\email{sutaoyu@iie.ac.cn}
\orcid{0009-0003-1674-7635}

\author{Xixun Lin}
\affiliation{%
    \institution{Institute of Information Engineering, \\ Chinese Academy of Sciences} 
    \state{Beijing}
    \country{China}
}
\email{linxixun@iie.ac.cn}
\orcid{0009-0004-6645-0597}

\author{Xiaodong Li}
\affiliation{%
    \institution{Institute of Information Engineering, \\ Chinese Academy of Sciences\\ School of Cyber Security, UCAS} 
    \state{Beijing}
    \country{China}
}
\email{lixiaodong@iie.ac.cn}
\orcid{0009-0008-7374-5413}

\author{Tingwen Liu}
\affiliation{%
    \institution{Institute of Information Engineering, \\ Chinese Academy of Sciences \\ School of Cyber Security, UCAS} 
    \state{Beijing}
    \country{China}
}
\email{liutingwen@iie.ac.cn}
\orcid{0000-0002-0750-6923}

\renewcommand{\shortauthors}{Jiawei Sheng et al.}

\begin{abstract}
Multi-domain knowledge graph completion (MKGC) aims to improve missing triple prediction in a target KG by transferring knowledge from other support KGs.
Existing methods typically enforce consistency constraints on equivalent entities across KGs to transfer knowledge, which risks suppressing domain-specific contextual information of entities.
This design can also compromise entity representation information from all KG domains, impeding performance improvements, especially in low-resource data scenarios. 
To address this, we pioneer a generation-based paradigm for MKGC and propose DMKGC, a conditional diffusion-guided knowledge transfer framework.
Our key insight is to treat each KG as a partial view of the entity entire information, and generate informative domain-general entity embeddings through diffusion models conditioned on support KGs.
Particularly, we first initialize domain-agnostic entity embeddings as prior entity embeddings, and then encode them within individual KGs.
Afterward, we fuse equivalent entities from support KGs as the conditional diffusion generation guidance. 
We leverage the prior entity embeddings as the proxy generation objective, which ensures this conditional generation to be unbiased towards any conditioned KGs.
Simultaneously, we also train the generated embeddings to be predictive across KGs, thus preserving domain-specific information.
Extensive experiments on 14 KGs in 3 benchmarks demonstrate a 4.3\% average MRR improvement in tail entity prediction over state-of-the-art methods, with sustained gains in low-resource data settings.
\end{abstract}



\maketitle

\section{Introduction}
Knowledge graphs (KGs), which structure knowledge as (head, relation, tail) triples, serve as a critical backbone for numerous web applications~\cite{KGQA:WWW23,KGRS:WWW25,KAG:WWW25}. 
However, their practical utility is often hampered by inherent incompleteness, prompting the task of KG completion (KGC)~\cite{tkde_survey_llmkg_pan, wang2017:kge}.
In general, KGC aims to infer missing elements within triples, typically predicting a missing tail entity given a head entity and a relation. 
This task can be challenging in scenarios with scarce data, where the limited observed triples severely hinder predictive performance~\cite{FSKGC:TWEB24,FAAN:EMNLP20}.

\begin{figure}[t]
	\begin{center}
        \includegraphics[width=0.98\columnwidth]{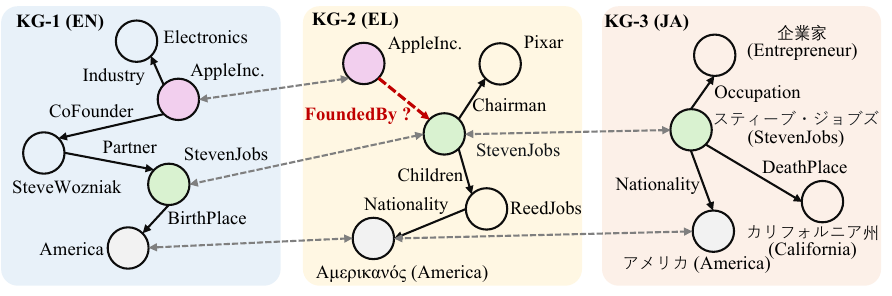}
		\caption{A toy example of the MKGC task, which predicts missing triples in the target KG with support KGs.}
		\label{Fig:Intro}
	\end{center}
\end{figure}

In recent years, numerous KGs have been constructed in different domains, which provide complementary knowledge and are promising for improving KGC. 
To this end, this paper focuses on \textit{multi-domain KG completion (MKGC)}\footnote{Here, we use the term \textit{domain} to generally denote a KG constructed from various sources, such as different languages or platforms~\cite{Chen2017:MTransE,ICML23:multisource,MDKGP:24,COLING25:DAEA}.}, a practical task that aims to predict missing triples in a \textit{target KG} by using other related \textit{support KGs}.
As shown in Figure~\ref{Fig:Intro}, the task is to predict (\texttt{Microsoft}, \texttt{Founder}, ?) in the target KG (EL), but the contextual triples of \texttt{BillGates} are sparse for prediction.
With the related triples in the support KG-1 (EN), the queried triple can be correctly inferred.
Here, the entities that appear simultaneously among multiple KGs (e.g. \texttt{BillGates}) are called \textit{equivalent entities}, which are previously aligned and connect across these KGs~\cite{Huang2022:SS-AGA,Tang2023:LSMGA}.

\begin{figure}[t]
	\begin{center}
            \includegraphics[width=0.98\columnwidth]{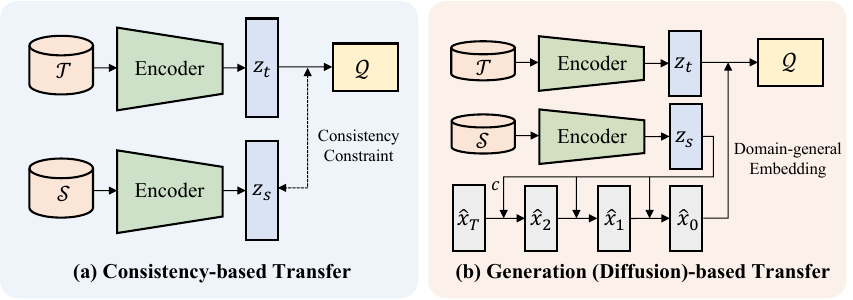}
		\caption{Paradigm of consistency-based and our generation-based knowledge transfer. 
        Here, $\mathcal{S}, \mathcal{T}$ are the support and target KG, $\mathcal{Q}$ is the query to predict new triples.
        }
		\label{Fig:Intro2}
	\end{center}
\end{figure}

Although the MKGC task is practical, it remains underexplored.
A core challenge lies in designing effective knowledge transfer modules to transfer valuable knowledge from support KGs to a target KG. 
Most existing studies~\cite{KEnS:chen-etal-2020-multilingual,CG_MuAlign,Singh2021:AlignKGC,Huang2022:SS-AGA,Tang2023:LSMGA,GLMKGC} follow a \textit{consistency-based} paradigm for solutions.
They learn entity embeddings within individual KGs, and then enforce consistency constraints~\cite{Chen2017:MTransE,Huang2022:SS-AGA,GLMKGC} between equivalent entities to ensure semantic alignment of KGs (shown in Figure~\ref{Fig:Intro2}(a)).
Despite their success, they mostly focus on the consistent information between KGs, neglecting entities' diverse contextual information.
This enforcement can potentially overshadow unique domain-specific information in each KG, which can overly regularize entity embeddings and limit predictability, especially in low-resource scenarios (shown in Section~\ref{Sec:Exp_Low_Res}).

To overcome this limitation, we propose a \textit{generation-based} knowledge transfer paradigm (Figure~\ref{Fig:Intro2}(b)).
The core idea is to learn to generate \textit{domain-general entity embeddings} that effectively preserve rich domain-specific information while remaining unbiased across domains.
To achieve this, we employ diffusion models (DMs)~\cite{DDPM:NIPS2020}, which generate such embeddings conditioned on the entity information from the support KGs.
Here, each KG can be seen as a partial observational view of the entire underlying entity.
This can be analogous to text-conditioned image generation, where different texts can guide the generation of the same entire visual concept~\cite{CFG:22,SDXL:ICLR2024}.
In contrast to existing consistency-based methods~\cite{Chen2017:MTransE,Huang2022:SS-AGA,GLMKGC,Tang2023:LSMGA}, our method generates a unified general-purpose embedding for each entity across domains, thereby avoiding overly rigid consistency constraints and yielding a more informative representation.

However, a pivotal challenge still remains: there are no real domain-general entity embeddings available to supervise the generation process.
To address this, we instead propose using domain-agnostic \textit{prior entity embeddings} as a proxy generation objective during training.
Since these priors are independent of any specific KG domains, they serve as an unbiased reference to guide the generation, ensuring that the resulting embeddings do not favor any particular conditioned KG.
Furthermore, to retain domain-specific information, we explicitly encourage the generated embeddings to be predictively effective within each KG, thereby preserving the domain-specific information presented in individual KGs.

Following the above idea, we propose DMKGC, a conditional \underline{D}iffusion guided knowledge transfer framework for \underline{MKGC}.
Specifically, given a target KG and multiple support KGs, we first share the initial entity embeddings as the prior entity embeddings, which are then independently encoded within the contextual structure of each individual KG.
Subsequently, we introduce a conditional diffusion model to generate domain-general entity embeddings, conditioned on the encoded representations from the support KGs.
These prior entity embeddings (prior to KG encoding) are reused to ensure unbiased generation, and the generated embeddings are also trained to remain KGC task-predictive in the target KG.
To realize the diffusion process, we design an attentive conditional denoiser that adaptively fuses the support KGs according to the target KG. 
In addition, we further devise a single-domain conditional regularization to enhance the generation stability, which treats each KG as an independent conditional guidance to generate a consistent objective.
Our major contributions can be summarized as follows:
\begin{itemize}[leftmargin=*]
    \item We pioneerly formulate MKGC in a generation-based manner, producing more informative domain-general representations.
    \item We propose a novel DMKGC\footnote{Our code is available at \url{https://github.com/JiaweiSheng/DMKGC}.} framework with DMs.
     It leverages support KGs as a condition, and simultaneously achieves task-predictive and domain-general information with constraints.
    \item Extensive experiments with 14 KGs in 3 benchmarks indicate significant 4.3\% averaged MRR improvements and show sustained improvements in low-resource data scenarios.
\end{itemize}

\section{Preliminaries}
In this section, we introduce the MKGC problem setup and the background of diffusion models.
\subsection{Problem Formulation}
Formally, let $\mathcal{D}=\{\mathcal{G}_i\}_{i=1}^N$ be a set of KGs, where each KG $\mathcal{G}_i=(\mathcal{E}_i,\mathcal{R}_i,\mathcal{F}_i)$ involves entities $\mathcal{E}_i$, relations $\mathcal{R}_i$ and factual triples $\mathcal{F}_i$.
For any two KGs $\mathcal{G}_i$ and $\mathcal{G}_j$, a small set of equivalent entity pairs is given: $\mathcal{A}_{ij}=\{(e_i, e_j)\mid e_i\equiv e_j, e_i\in\mathcal{E}_i,\ e_j\in\mathcal{E}_j\}$. 
Additionally, all KGs adhere to a unified relation schema $\mathcal{R}$, meaning each $\mathcal{R}_i\subseteq\mathcal{R}$. 
The task is to predict query triples $\mathcal{Q}\!=\!\{( h, r, ?)\}$ for a target KG $\mathcal{T}=\mathcal{G}_t,\mathcal{G}_t\in \mathcal{D}$, leveraging existing triples from both the target KG $\mathcal{T}$ and all other support KGs $\mathcal{S}=\{\mathcal{G}_s|\mathcal{G}_s\in\mathcal{D},s\neq t\}$.

\subsection{Diffusion Model}\label{Sec:DMs}
In this paper, we leverage diffusion models (DMs)~\cite{DDPM:NIPS2020,DULNT} to transfer knowledge, which contains a forward and a reverse process.

\begin{figure*}[!t]
\centering
\includegraphics[width=\textwidth]{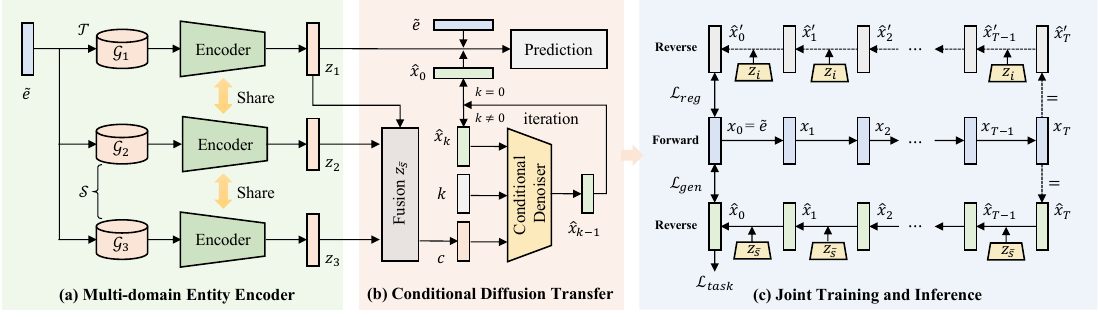}
\caption{The overview of our proposed framework, DMKGC. It contains (a) \textit{multi-domain entity encoder} to extract KG features, (b) \textit{conditional diffusion transfer} to generate entity embeddings, and (c) \textit{joint training and inference} to achieve the task.}
\label{Fig:DMKGC}
\end{figure*}

\subsubsection{Forward Process}
Given an input data sample $\bm{x}_0\sim q(x_0)$, the forward process constructs a series of latent variables $\bm{x}_{1:T}$ through a Markov chain by gradually adding Gaussian noise across $T$ steps. 
Specifically, the transition at each time step $k$ is defined by:
\begin{equation}
\begin{aligned}
q(x_{k} | \bm{x}_{k-1})=\mathcal{N}(\bm{x}_{k} ; \sqrt{1-\beta_{k}} \bm{x}_{k-1}, \beta_{k} \bm{I}),
\end{aligned}
\end{equation}
where $k\in \{1,2,...,T\}$, $\mathcal{N}$ denotes a Gaussian distribution, and factor $\beta_{k}\in(0, 1)$ controls the scale of noise added at step $k$. 
The factors ensure that, when $T\rightarrow \infty$, the variable $\bm{x}_{k}$ converges to a standard Gaussian, allowing sampling from a Gaussian noise to generate real samples in the reverse process.

\subsubsection{Reverse Process}
DMs learn to iteratively reconstruct data by reversing the forward process trajectory.
Formally, starting from an initial state $\bm{x}_T\sim \mathcal{N}(0,1)$, DMs parameterize a Markov chain that transitions from $\bm{x}_k$ to $\bm{x}_{k-1}$ via:
\begin{equation}
\begin{aligned}
p_{\theta}(\bm{x}_{k-1} | \bm{x}_{k})=\mathcal{N}(\bm{x}_{k-1} ; \bm{\mu}_{\theta}(\bm{x}_{k}, k), \Sigma_{\theta}(\bm{x}_{k}, k)),
\end{aligned}
\end{equation} 
where $k\in \{T, T-1, ..., 1\}$ denotes the reversed time step, $\bm{\mu}_\theta(\bm{x}_k, k)$ and $\bm{\Sigma}_\theta(\bm{x}_k, k)$ are the mean and covariance of the Gaussian distribution generated by a neural network $\theta$.
For simplicity and training stability, $\Sigma_{\theta}(\bm{x}_{k}, k)$ is usually set to constants $\sigma^2(k)\bm{I}$ varying over time steps~\cite{DDPM:NIPS2020}.
In this way, the reverse process can be seen as a step-by-step denoising process capturing minor changes.

\subsubsection{Training}
The objective function is optimized by maximizing the Evidence Lower Bound (ELBO) of the likelihood of the observed data $\bm{x}_0$ to enable data generation~\cite{DDPM:NIPS2020}, which is:
\begin{equation}\label{Eq:Diff_ELBO}
\begin{aligned}
\log p(\bm{x}_{0}) &=\log\int p(\bm{x}_{0: T})\mathrm{d}\bm{x}_{1: T} =\log\mathbb{E}_{\bm{q}(\bm{x}_{1: T}|\bm{x}_{0})}[\frac{p(\bm{x}_{0: T})}{q(\bm{x}_{1: T}|\bm{x}_{0})}]\\
&\geq\underbrace{\mathbb{E}_{\bm{q}(\bm{x}_{1}|\bm{x}_{0})}[\log p_{\theta}(\bm{x}_{0}|\bm{x}_{1})]}_{\text{reconstruction term}\ \mathcal{L}_0}-\underbrace{D_{\mathrm{KL}}(q(\bm{x}_{T}|\bm{x}_{0})\| p(\bm{x}_{T}))}_{\text{prior matching term}\ \mathcal{L}_{T}}\\
&\quad -\sum_{k=2}^{T}\underbrace{\mathbb{E}_{\bm{q}(\bm{x}_{k}|\bm{x}_{0})}[D_{\mathrm{KL}}(q(\bm{x}_{k-1}|\bm{x}_{k},\bm{x}_{0})\| p_{\theta}(\bm{x}_{k-1}|\bm{x}_{k}))]}_{\text{denoising matching term}\ \mathcal{L}_{k-1}},
\end{aligned}
\end{equation}
where there are three terms to resolve:
\begin{itemize}[leftmargin=*]
    \item \textit{Reconstruction term}: Measures fidelity of $\bm{x}_0$ given state $\bm{x}_1$, which can be achieved by mean square error (MSE).
    \item \textit{Prior matching term}: Ensures $q(\bm{x}_T|\bm{x}_0)$ converges to the Gaussian prior $p(\bm{x}_T)\sim \mathcal{N}(\bm{0},\bm{I})$, which is constant with no trainable parameters and usually omitted in optimization.
    \item \textit{Denoising matching terms}: Minimizes the Kullback-Leibler (KL) divergence between the true probability $q(\bm{x}_{k-1}|\bm{x}_{k},\bm{x}_{0})$ (analytically tractable) and the learned probability $p_{\theta}(\bm{x}_{k-1}|\bm{x}_{k})$, allowing iterative generation of $\bm{x}_{k-1}$ from $\bm{x}_{k}$. By simplification~\cite{DDPM:NIPS2020}, this term can be achieved by $\sum_{k=2}^{T} \mathbb{E}_{k, \bm{\epsilon}} [ \| \bm{\epsilon} - \bm{\epsilon}_{\theta} ( \bm{x}_{k}, k ) \|_{2}^{2} ]$, where $\bm{\epsilon}_{\theta} (\bm{x}_{k}, k)$ is a neural denoiser (e.g., U-Net~\cite{DDPM:NIPS2020}, MLP~\cite{DiffusionRM}) that predicts the added noise $\bm{\epsilon}$ at time step $k$.
\end{itemize}

\subsubsection{Inference \& Sampling}
With the trained neural denoiser $\theta$, the DMs can sample $\bm{x}_T\sim \mathcal{N}(\bm{0},\bm{I})$ and iteratively leverage $p_{\theta}(\bm{x}_{k-1}|\bm{x}_{k})$ to generate the predicted sample $\hat{\bm{x}}_0$.
Note that all above procedures can be achieved by embeddings for efficiency~\cite{LDM:CVPR2022}.

\section{Methodology}
Our core idea is to generate domain-general entity embeddings that transfer knowledge from support KGs to enhance the target KG in KGC prediction.
The framework is presented in Figure~\ref{Fig:DMKGC}.

\subsection{Multi-domain Entity Encoder}

\subsubsection{Prior Entity Embedding}
Prior to KG domain encoding, we obtain entity embeddings via a randomly initialized layer as:
\begin{equation}\label{Eq:unbiased_emb}
\begin{aligned}
\tilde{\bm{e}} = \mathrm{Embedding}(e), e\in \mathcal{E},
\end{aligned}
\end{equation}
where we call $\tilde{\bm{e}} \in \mathbb{R}^d$ the \textit{prior entity embedding}.
Here, we share the embeddings for all KGs, where $\mathcal{E}$ denotes the unified set of entities.
Note that these embeddings are domain-independent and obtained prior to domain-specific encoding, making them inherently unbiased toward any domains.

\subsubsection{Domain Entity Embedding}
To encode KG domain information, we leverage an effective KG encoder~\cite{Tang2023:LSMGA} .
Given a KG $\mathcal{G}$ (omit subscript for simplicity), it encodes an entity $e$ with its relational neighborhood $\mathcal{N}(e)$ from $\mathcal{G}$ as:
\begin{equation}\label{Eq:Encoder}
\begin{aligned}
\mathrm{Encoder}_l(\bm{e},\mathcal{G}) &= \bm{e} + \delta(\!\!\!\!\sum_{\{r_{j}, e_{j}\}\in \mathcal{N}(e)} \!\!\!\!\alpha_l(\bm{e}, \bm{r}_{j}, \bm{e}_{j})\cdot \bm{W}^l_{3}[\bm{e}_{j}\oplus \bm{r}_{j}]), \\
\alpha_l(\bm{e}, \bm{r}_{j}, \bm{e}_{j}) &= \operatorname{softmax}(\mathrm{score}_l(\bm{e}, \bm{r}_{j}, \bm{e}_{j})),\\
\mathrm{score}_l(\bm{e}, \bm{r}_{j}, \bm{e}_{j}) &= \beta_{r_{j}} \cdot \frac{1}{\sqrt{d}}(\bm{W}^l_{1}\bm{e}\cdot\bm{W}^l_{2}[\bm{e}_{j}\oplus\bm{r}_{j}]),
\end{aligned}
\end{equation}
where $\delta$ is the ReLU activation, $\oplus$ is the vector concatenation, $l$ is the layer, $\bm{W}^l_1\in \mathbb{R}^{d\times d}, \bm{W}^l_2,\bm{W}^l_3\in \mathbb{R}^{d\times 2d}$ are learnable layer weights. 
The relation embedding $\bm{r}\in \mathbb{R}^{d}$ is randomly initialized, and the input $\bm{e}$ is the prior entity embedding $\tilde{\bm{e}}$ from Eq.~(\ref{Eq:unbiased_emb}).
Here, the score considers both the learnable prior weight $\beta_{r}\in \mathbb{R}$ of relation $r$, and also captures the contextual attentive relational relevance.
The encoder output is denoted as $\bm{z}$, which we call \textit{domain entity embeddings} as they contain domain-specific information.

In the following sections, we use $\bm{z}_t$ to denote an involved entity $e$ from the target KG $\mathcal{T}=\mathcal{G}_t$, and $\{\bm{z}_s\}_{s}$ denotes its equivalent entities\footnote{Following~\citet{Tang2023:LSMGA}, for simplicity of implementation, we also add virtual isolated entities in the KGs where the equivalent entity doesn't exist.} from the corresponding support KGs $\mathcal{S}=\{\mathcal{G}_s\}_{s}$.

\subsection{Conditional Diffusion Transfer}
Unlike existing consistency-based methods~\cite{Chen2017:MTransE,Huang2022:SS-AGA,GLMKGC}, we transfer knowledge through conditional diffusion.
To this end, we leverage support KGs as conditional guidance and use the prior entity embeddings as the proxy generation objective to ensure an unbiased generation.
The training will be detailed in Section~\ref{Sec:Training&Inference}.

\subsubsection{Forward Process}
Given an entity $e$ and its domain entity embeddings, we expect to generate a domain-general entity embedding.
To this end, we use the prior embedding $\tilde{\bm{e}}$ as the initial state, i.e., $\bm{x}_0=\tilde{\bm{e}}$.
As introduced in section~\ref{Sec:DMs}, the forward process is $q(x_{k} | \bm{x}_{k-1})=\mathcal{N}(\bm{x}_{k} ; \sqrt{1-\beta_{k}} \bm{x}_{k-1}, \beta_{k} \bm{I})$, where $\beta_k$ is the noise scale factor.
Using the reparameterization trick~\cite{vae,DDPM:NIPS2020}, we can efficiently obtain $\bm{x}_k$ from $\bm{x}_0$ as:
\begin{equation}
\begin{aligned}\label{Eq:forwardT}
q(\bm{x}_{k}| \bm{x}_{0})=\mathcal{N}(\bm{x}_{k} ; \sqrt{\bar{\alpha}_{k}} \bm{x}_{0},(1-\bar{\alpha}_{k}) \bm{I}),
\end{aligned}
\end{equation}
where $\alpha_k=1-\beta_k$, $\bar{\alpha}_k=\Pi_{k'=1}^k\alpha_k$.
Hence, we can directly obtain the embedding $\bm{x}_k=\sqrt{\bar{\alpha}_k}\bm{x}_0 +\sqrt{1-\bar{\alpha}_k}\epsilon, \epsilon\sim \mathcal{N}(\bm{0,\bm{I}})$ at any step $k$.
The last state of the noisy embedding is denoted as ${\bm{x}}_T$.

\subsubsection{Reverse Process}
The reverse process gradually denoises $\bm{x}_T$ to reconstruct the initial state $\bm{x}_0$.
Believing that each KG describes a partial view of the domain-general entity, we leverage the equivalent entities as condition information for reconstruction. 
Thereafter, the reverse process with conditions~\cite{CFG:22} is as follows:
\begin{equation}
\begin{aligned}
p_{\theta}(\bm{x}_{k-1} | \bm{x}_{k}, \bm{c})=\mathcal{N}(\bm{x}_{k-1} ; \bm{\mu}_{\theta}(\bm{x}_{k}, \bm{c}, k), \sigma^2(k)\bm{I}),
\end{aligned}
\end{equation}
where $\bm{\mu}_{\theta}(\bm{x}_{k}, \bm{c}, k)$ is learned by a conditional denoiser (detailed later).
Here, $\bm{c}\in \mathbb{R}^d$ is the embedding of one or more equivalent entities to guide the generation (i.e., noise prediction).

\subsubsection{Conditional Denoiser}
The conditional denoiser learns to predict the added noise at each time step $k$.
To keep it simple, we employ an effective multilayer perceptron (MLP) as the denoiser:
\begin{equation}\label{Eq:MLP_Denoiser}
\begin{aligned}
\hat{\bm{x}}_{\theta}(\bm{x}_k,\bm{c},k) := \operatorname{MLP}(\bm{x}_k\oplus \bm{c}\oplus\bm{k};\theta),
\end{aligned}
\end{equation}
where we use a two-layer MLP with dimensions as $3d \rightarrow 2d\rightarrow d$, and $\bm{k}\in \mathbb{R}^{d}$ is the time-step embedding~\cite{DDPM:NIPS2020}.
Here, as the condition information can be different, we consider three types of condition $\bm{c}$:
(i) when the condition is not given, we set $\bm{c}=\phi$ where $\phi\in \mathbb{R}^d$ is an initialized \textit{null} embedding in training.
(ii) when an equivalent entity embedding from an arbitrary KG $\mathcal{G}_i$ is given, the condition is $\bm{c}=\bm{z}_i$.
(iii) when equivalent entities $\{\bm{z}_s\}_s$ from support KGs $\mathcal{S}$ are given, we fuse them according to their relevance to the entity $\bm{z}_t$ as
\begin{equation}\label{Eq:Att_cond}
\begin{aligned}
\bm{c}=\bm{z}_{\bar{s}} &= \sum_{\bm{z}_s\in \{\bm{z}_s\}_s}\alpha(\bm{z}_{t},\bm{z}_{s}) \cdot \bm{z}_s, \\
\alpha(\bm{z}_{t},\bm{z}_{s}) &= \operatorname{softmax}(\mathrm{score}(\bm{z}_{t},\bm{z}_{s})),\\
\mathrm{score}(\bm{z}_{t},\bm{z}_{s}) &= \beta_{ts} \cdot \frac{1}{\sqrt{d}}(\bm{U}_{1}\bm{z}_{t}\cdot\bm{U}_{2}\bm{z}_{s}]),
\end{aligned}
\end{equation}
where $\bm{U}_{1},\bm{U}_{2}\in \mathbb{R}^{d\times d}$ are learnable weights.
In particular, $\beta_{ts}\in \mathbb{R}$ is a learnable factor to reflect the prior relevance between $\mathcal{G}_t$ and $\mathcal{G}_s$.
This design captures both the prior and contextual relevance of equivalent entities between KGs, aggregating conditions adaptive to the target KG to benefit generation.

\subsection{Joint Training and Inference}\label{Sec:Training&Inference}

\subsubsection{Training}
Conventional DMs train the denoiser to predict the added noise at each time step~\cite{DDPM:NIPS2020}.
However, it is costly to perform a reverse process in training to obtain the generated domain-general embedding.
Hence, we look back on ELBO in Eq.~(\ref{Eq:Diff_ELBO}).

Consider the denoising matching term $\mathcal{L}_{k-1}$, which matches $p_\theta(\bm{x}_{k-1}|\bm{x}_k)$ with $q(\bm{x}_{k-1}|\bm{x}_k, \bm{x}_0)$ by KL divergence.
By Bayes rules, $q(\bm{x}_{k-1}|\bm{x}_k, \bm{x}_0)\!\sim\! \mathcal{N}(\bm{x}_{k-1}; \bm{\tilde{\mu}}(\bm{x}_{k}, \bm{x}_{0}, k), \sigma^{2}(k) \bm{I})$ can be written as:
\begin{equation}
\begin{aligned}
\bm{\tilde{\mu}}(\bm{x}_{k}, \bm{x}_{0}, k)&=\frac{\sqrt{\alpha_{k}}(1-\bar{\alpha}_{k-1})}{1-\bar{\alpha}_{k}} \bm{x}_{k}+\frac{\sqrt{\bar{\alpha}_{k-1}}(1-\alpha_{k})}{1-\bar{\alpha}_{k}} \bm{x}_{0}, \\
\sigma^{2}(k)\bm{I}&=\frac{(1-\alpha_{k})(1-\bar{\alpha}_{k-1})}{1-\bar{\alpha}_{k}}\bm{I}.
\end{aligned}
\end{equation}
where $\tilde{\mu}(\bm{x}_{k}, \bm{x}_{0}, k)$ and $\sigma^{2}(k)\bm{I}$ are mean and covariance.
As suggested~\cite{DreamRec:NeurIPS2023,DiffusionRM}, instead of learning the added noise $\bm{\epsilon}$ by parameterizing $\bm{\epsilon}_\theta(\bm{x}_k, k)$, we directly parameterize $\bm{x}_0$, which is:
\begin{equation}\label{Eq:mu_theta}
\bm{\mu}_{\theta}(\bm{x}_{k}, k)=\frac{\sqrt{\alpha_{k}}(1-\bar{\alpha}_{k-1})}{1-\bar{\alpha}_{k}} \bm{x}_{k}+\frac{\sqrt{\bar{\alpha}_{k-1}}(1-\alpha_{k})}{1-\bar{\alpha}_{k}} \hat{\bm{x}}_{\theta}(\bm{x}_{k}, k),
\end{equation}
where $\hat{\bm{x}}_{\theta}(\bm{x}_{k}, k)$ is an approximation of the given true data $\bm{x}_0$, based on the situation at the time step $k$.
This is proven to be equal to the vanilla ELBO in DDPM~\cite{DreamRec:NeurIPS2023}.
Using the above parameterization, the \textit{denoising matching term} ($k\geq 2$) can be derived as:
\begin{equation}\label{Eq:L_{k-1}}
\begin{aligned}
\mathcal{L}_{k-1} &:= \mathbb{E}_{q(\bm{x}_{k} | \bm{x}_{0})}[D_{\mathrm{KL}}(q(\bm{x}_{k-1} | \bm{x}_{k}, \bm{x}_{0}) \| p_{\theta}(\bm{x}_{k-1} | \bm{x}_{k}))] \\
&= \mathbb{E}_{q(\bm{x}_{k} | \bm{x}_{0})}[\frac{1}{2 \sigma^{2}(k)}\|\bm{\mu}_{\theta}(\bm{x}_{k}, k)-\tilde{\bm{\mu}}(\bm{x}_{k}, \bm{x}_{0}, k)\|_{2}^{2}] \\
&=\mathbb{E}_{q(\bm{x}_{k} | \bm{x}_{0})}[\frac{1}{2}(\frac{\bar{\alpha}_{k-1}}{1-\bar{\alpha}_{k-1}}-\frac{\bar{\alpha}_{k}}{1-\bar{\alpha}_{k}})\|\hat{\bm{x}}_{\theta}(\bm{x}_{k}, k)-\bm{x}_{0}\|_{2}^{2}].
\end{aligned}
\end{equation}
This form is similar to the \textit{reconstruction term} ($k=0$) with Gaussian log-likelihood~\cite{MultiVAE:WWW18} as $\mathcal{L}_{0} :=  \mathbb{E}_{q(\bm{x}_{1} | \bm{x}_{0})}\!\|\hat{\bm{x}}_{\theta}(\bm{x}_{1}, 1) - \bm{x}_{0}\|_{2}^{2}$.
Thereafter, we combine them into a \textit{unified training term} ($k\geq 1$) as:
\begin{equation}\label{Eq:L_unified}
\begin{aligned}
\mathcal{L}_{k-1} := \mathbb{E}_{q(\bm{x}_{k} | \bm{x}_{0})}\|\hat{\bm{x}}_{\theta}(\bm{x}_{k}, k)-\bm{x}_{0}\|_{2}^{2}.
\end{aligned}
\end{equation}
In this way, the denoiser actually seeks to predict the initial embedding (i.e., the generation objective) at each step.
Based on this design, it provides a direct way to impose constraints to manipulate the generated embeddings.
Here, we propose three constraints:

\paragraph{(i) Domain-General Embedding Generation}
To generate domain-general entity embeddings, we leverage all equivalent entities from support KGs to guide generation, which is
\begin{equation}\label{Eq:L_gen}
\begin{aligned}
\mathcal{L}_{\text{gen}} := \sum_{k=1}^{T} \mathbb{E}_{q(\bm{x}_{k} | \tilde{\bm{e}})}\|\hat{\bm{x}}_{\theta}(\bm{x}_{k},\bm{z}_{\bar{s}}, k)-\tilde{\bm{e}}\|_{2}^{2},
\end{aligned}
\end{equation}
where $\hat{\bm{x}}_{\theta}(\bm{x}_{k},\bm{z}_{\bar{s}}, k)$ comes from Eq.~(\ref{Eq:Att_cond}).
This constraint encourages the generated embeddings to approximate the prior embeddings, ensuring unbiased generation towards conditioned support KGs.
Here, we adopt the classifier-free guidance (CFG) strategy~\cite{CFG:22}, which retains a ratio $p_{u}$ of generation cases without conditions (i.e., $\phi$), improving the unconditional generation ability.

\paragraph{(ii) Target-Domain Task Prediction}
This constraint ensures the generated domain-general embedding to be task-predictive in the KG target $\mathcal{G}_t$.
To enrich the entity information, we fuse the generated $\hat{\bm{x}}_{\theta}$, the prior $\tilde{\bm{e}}$, and the target-domain embedding $\bm{z}_{t}$ as:
\begin{equation}\label{Eq:entity_embedding_combine}
\begin{aligned}
\bar{\bm{z}}_{t} = \hat{\bm{x}}_{\theta} + \tilde{\bm{e}} + \bm{z}_{t},
\end{aligned}
\end{equation}
where $\hat{\bm{x}}_{\theta}=\hat{\bm{x}}_{\theta}(\bm{x}_{k},\bm{z}_{\bar{s}}, k)$.
For a triple $(h,r,o)$ sampled\footnote{Here, we use $o$ to denote the tail entity, since $t$ is used to denote the target KG.} in $\mathcal{G}_t$, we use the classical triple scoring function~\cite{transE} to predict plausibility:
\begin{equation}\label{Eq:TripleScore}
\psi(h, r, o) = -\|\bar{\bm{z}}_{t,h} + \bm{r} - \bar{\bm{z}}_{t,o}\|_{2},
\end{equation}
\begin{equation}\label{Eq:L_task}
\mathcal{L}_{\text{task}} :=\!\!\! \sum_{(h, r, o)\in\mathcal{F}_t} \sum_{(h, r,o')\notin\mathcal{F}_t}\!\!\![\lambda - \psi(h, r, o) + \psi(h, r, o')]_{+},
\end{equation}
where we adopt the margin loss, $[\cdot]_+:=\max(\cdot,0)$, $\lambda$ is the margin factor, and $o'\in \mathcal{E}_t$ is a negative entity randomly selected. 
In this way, the generated embedding $\hat{\bm{x}}_{\theta}$ is refined, and together with other embeddings to enhance KGC in the target KG.

\paragraph{(iii) Single-Domain Conditional Regularization}
This constraint further emphasizes the unbiased and consistent generation with partial conditions.
Specifically, it seeks to generate consistent domain-general embedding conditioned on each KG:
\begin{equation}\label{Eq:L_reg}
\begin{aligned}
\mathcal{L}_{\text{reg}} := \sum_{i=1}^{N}\sum_{k=1}^{T} \mathbb{E}_{q(\bm{x}_{k} | \tilde{\bm{e}})}\|\hat{\bm{x}}_{\theta}(\bm{x}_{k},\bm{z}_i, k)-\tilde{\bm{e}}\|_{2}^{2},
\end{aligned}
\end{equation}
where $\bm{z}_i$ is a domain entity embedding of equivalent entities from all KGs.
This further ensures generation consistency in practice.

Based upon the aforementioned constraints, the overall training objective is formulated as:
\begin{equation}\label{Eq:Overall_loss}
\begin{aligned}
\mathcal{L} = \mathcal{L}_{\text{task}}+\omega_1\mathcal{L}_{\text{gen}}+\omega_2\mathcal{L}_{\text{reg}},
\end{aligned}
\end{equation}
where $\omega_1,\omega_2\in\mathbb{R}$ are harmonic factors to balance training.
The algorithm~\ref{Alg:training} presents the overall training procedure, where we adopt sampling strategies~\cite{DDPM:NIPS2020} to accelerate training for efficiency.

\subsubsection{Inference}
In inference, we use the domain-general entity embedding $\hat{\bm{x}}_0$ from $\hat{\bm{x}}_\theta(\bm{x}_k,\bm{z}_{\bar{s}},k)$ to transfer knowledge from the support KGs.
Following the CFG strategy~\cite{CFG:22} to enable conditional generation, we adjust the generation by interpolating the generated conditional and unconditional embedding as:
\begin{equation}\label{Eq:cfg_infer}
\begin{aligned}
\tilde{\bm{x}}_\theta(\bm{x}_k,\bm{z}_{\bar{s}},k) = \hat{\bm{x}}_\theta(\bm{x}_k,\phi, k) + s[\hat{\bm{x}}_\theta(\bm{x}_k,\bm{z}_{\bar{s}},k)- \hat{\bm{x}}_\theta(\bm{x}_k,\phi,k)],
\end{aligned}
\end{equation}
where $s\in \mathbb{R}$ is a factor in controlling the strength of the condition.
Subsequently, in inference, given $\bm{x}_0=\tilde{\bm{e}}$, we first perform the forward process to derive $\bm{x}_T$, and then set $\hat{\bm{x}}_{T}=\bm{x}_{T}$ to start the reverse process.
This process holds for both head and tail entities involved (as in Eq.~(\ref{Eq:TripleScore})).
For a query $(h,r,?)$ in $\mathcal{G}_t$, we treat all entities in $\mathcal{E}_t$ as candidates.
The entity with the highest score is returned as the result.
The algorithm~\ref{Alg:inference} presents the inference procedure.

\begin{algorithm}[!t]
\caption{Training procedure}
\small
\label{Alg:training}
\hbox{\textbf{Input}: Training data $\mathcal{D}=\{\mathcal{G}_i\}_i$ and equivalent entity sets $\{\mathcal{A}_{ij}\}_{i\neq j}$.}
\hbox{\textbf{Output}: Model parameters $\Theta$.}
\begin{algorithmic}[1] 
\STATE Initialize all model parameters. \\
\STATE \textbf{while} not convergence \textbf{do} \\
\STATE \quad Sample a target KG $\mathcal{G}_t$ and support KGs $\mathcal{S}=\{\mathcal{G}_s|s\neq t\}$.
\STATE \quad Sample a triple $(h,r,o)$ from $\mathcal{G}_t$, and a negative entity $o'$ from $\mathcal{E}_t$.
\STATE \quad \textbf{for} $e \in \{h,o,o'\}$ \textbf{do}
\STATE \quad \quad Obtain prior embedding $\tilde{\bm{e}}$, and let $\bm{x}_0=\tilde{\bm{e}}$;
\STATE \quad \quad  \textbf{for} all $\mathcal{G}_i \in \mathcal{D}$ \textbf{do}
\STATE \quad \quad \quad Obtain $\bm{z}_i=\mathrm{Encoder}(\tilde{\bm{e}},\mathcal{G}_i)$ by Eq.~(\ref{Eq:Encoder});
\STATE \quad \quad \textbf{end for}
\STATE \quad  \quad Sample $k \sim \mathcal{U}(1, T)$, $\epsilon \sim \mathcal{N}(0, I)$, and obtain $\bm{x}_k$ by Eq.~(\ref{Eq:forwardT}); 
\STATE \quad \quad Estimate $\hat{\bm{x}}_\theta(\bm{x}_k,\bm{z}_{\bar{s}},k)$ with CFG, obtain $\mathcal{L}_\mathrm{gen}$ by Eq.~(\ref{Eq:L_gen});
\STATE \quad \quad Estimate $\hat{\bm{x}}_\theta(\bm{x}_k,\bm{z}_i,k)$ with CFG, obtain $\mathcal{L}_\mathrm{reg}$ by Eq.~(\ref{Eq:L_reg});
\STATE \quad \quad  Obtain mixed embedding $\bar{\bm{z}}_{t,e}=\hat{\bm{x}}_{\theta} + \tilde{\bm{e}} + \bm{z}_{t,e}$;
\STATE \quad \textbf{end for}
\STATE \quad  Obtain $\mathcal{L}_\mathrm{task}$ by Eq.~(\ref{Eq:L_task});
\STATE \quad Obtain overall loss $\mathcal{L}$ by Eq.~(\ref{Eq:Overall_loss}) to optimize $\Theta$;
\STATE\textbf{end while}
\end{algorithmic}
\end{algorithm}

\begin{algorithm}[!t]
\caption{Inference procedure}
\small
\label{Alg:inference}
\begin{algorithmic}[1]
\hbox{\textbf{Input}: $\Theta$, query $(h,r,?)$ in target $\mathcal{T}=\mathcal{G}_t$, support $\mathcal{S}=\{\mathcal{G}_s\}_{s\neq t}$.}
\hbox{\textbf{Output}: Score $\psi(h,r,o')$ of all candidate entities $o'\in \mathcal{E}_{t}$.}
\STATE \textbf{for} $e \in \{h\}\cup\mathcal{E}_t$ \textbf{do}
\STATE \quad Obtain prior embedding $\tilde{\bm{e}}$.
\STATE \quad  \textbf{for} all $\mathcal{G}_i \in \mathcal{T}\cup\mathcal{S}$ \textbf{do}
\STATE \quad\quad Obtain $\bm{z}_i=\mathrm{Encoder}(\tilde{\bm{e}}, \mathcal{G}_i)$ by Eq.~(\ref{Eq:Encoder}).
\STATE \quad \textbf{end for}
\STATE \quad Sample $\epsilon \sim \mathcal{N}(0, I)$, and obtain $\hat{\bm{x}}_T$ by Eq.~(\ref{Eq:forwardT}); 
\STATE \quad \textbf{for} $k=T, \cdots, 1$ \textbf{do}
\STATE \quad\quad Estimate $\hat{\bm{x}}_\theta(\hat{\bm{x}}_k,\bm{z}_{\bar{s}},k)$ and $\hat{\bm{x}}_\theta(\hat{\bm{x}}_k,\phi,k)$ by Eq.~(\ref{Eq:MLP_Denoiser});
\STATE \quad\quad Obtain $\tilde{\bm{x}}_\theta(\hat{\bm{x}}_k,\bm{z}_{\bar{s}},k)$ by Eq.~(\ref{Eq:cfg_infer});
\STATE \quad\quad Obtain $\hat{\bm{x}}_{k-1}=\mu_{\theta}(\hat{\bm{x}}_k,k)$ with $\hat{\bm{x}}_k$ and $\tilde{\bm{x}}_\theta(\hat{\bm{x}}_k,\bm{z}_{\bar{s}},k)$ by Eq.~(\ref{Eq:mu_theta});
\STATE \quad \textbf{end for}
\STATE \quad  Obtain mixed embedding $\bar{\bm{z}}_t=\hat{\bm{x}}_0 + \tilde{\bm{e}} + \bm{z}_{t}$;
\STATE \quad  Calculate score $\psi(h,r,o')$ with $\bar{\bm{z}}_{t,h}$ and $\bar{\bm{z}}_{t,o'}$ by Eq.~(\ref{Eq:TripleScore});
\STATE \textbf{end for}
\end{algorithmic}
\end{algorithm}

\section{Experiments}
In this section, we address the following research questions:
\textbf{(Q1)} How effectively does DMKGC perform across diverse benchmark datasets?
\textbf{(Q2)} Is the proposed knowledge transfer robust to low-resource data scenarios?
For analyzes of parameter sensitivity and computational efficiency, please refer to the \textbf{Appendix} \textbf{(Q3)}.

\subsection{Evaluation Settings}

\begin{table*}[!t]
\small 
\caption{Results (\%) on DBP-5L. $\dag$ indicates re-produced results. The best result is \textbf{bold-faced} and the runner-up is \underline{underlined}.}
\centering
\setlength{\tabcolsep}{3pt}
\begin{tabular*}{1 \textwidth}{@{\extracolsep{\fill}}@{}lccc|ccc|ccc|ccc|ccc|c@{}}
\toprule
\multirow{2.5}{*}{\bf Method} & \multicolumn{3}{c}{\bf EL} & \multicolumn{3}{c}{\bf EN} &  \multicolumn{3}{c}{\bf ES} & \multicolumn{3}{c}{\bf FR} & \multicolumn{3}{c}{\bf JA} &\bf AVG \\
\cmidrule{2-4}\cmidrule{5-7}\cmidrule{8-10} \cmidrule{11-13}\cmidrule{14-16}\cmidrule{17-17}
& H@1  & H@10   &  MRR  
& H@1  & H@10   &  MRR  
& H@1  & H@10   &  MRR     
& H@1  & H@10   &  MRR     
& H@1  & H@10   &  MRR & MRR
 \\
\midrule
TransE  & 13.1 & 43.7 & 24.3 & 7.3 & 29.3 & 16.9 & 13.5 & 45.0 & 24.4 & 17.5 & 48.8 & 27.6 & 21.1 & 48.5 & 25.3 & 23.7 \\
DistMult & 8.9 & 11.3 & 9.8 & 8.8 & 30.0 & 18.3 & 7.4 & 22.4 & 13.2 & 6.1 & 23.8 & 14.5 & 9.3 & 27.5 & 15.8 & 14.3 \\
RotatE  & 14.5 & 36.2 & 26.2 & 12.3 & 30.4 & 20.7 & 21.2 & 53.9 & 33.8 & 23.2 & 55.5 & 35.1 & 26.4 & 60.2 & 39.8  &  31.1 \\
KG-BERT & 17.3 & 40.1 & 27.3 & 12.9 & 31.9 & 21.0 & 21.9 & 54.1 & 34.0 & 23.5 & 55.9 & 35.4 & 26.9 & 59.8 & 38.7    & 31.3 \\
\midrule
KEnS & 28.1 & 56.9 & - & 15.1 & 39.8 & - & 23.6 & 60.1 & - & 25.5 & 62.9 & - & 32.1 & 65.3 & -   & - \\
CG-MuA & 21.5 & 44.8 & 32.8 & 13.1 & 33.5 & 22.2 & 22.3 & 55.4 & 34.3 & 24.2 & 57.1 & 36.1 & 27.3 & 61.1 & 40.1     & 33.1\\
AlignKGC & 27.6 & 56.3 & 33.8 & 15.5 & 39.2 & 22.3 & 24.2 & 60.9 & 35.1 & 24.1 & 62.3 & 37.4 & 31.6 & 64.3 & 41.6  & 34.0 \\
SS-AGA & 30.8 & 58.6 & 35.3 & 16.3 & 41.3 & 23.1 & 25.5 & 61.9 & 36.6 & 27.1 & 65.5 & 38.3 & {34.6} & 66.9 & 42.9  & 35.2 \\
LSMGA & 33.1  & \underline{89.9} & \underline{54.5} & {16.8} & \underline{61.7} & {32.4} & {25.6} & \underline{74.8} & {42.8} & {31.2} & \underline{81.3} & \underline{48.6} & 33.5 & \underline{79.1} & {49.8}  & 45.6 \\
GLKGC$^\dag$ & \underline{36.6} & 86.5 & 53.0 & \underline{17.1} & 60.2 & \underline{32.9} & \underline{28.3} & 74.4 & \underline{43.6} & \underline{31.5} & 78.4 & 47.9 & \underline{36.5} & 77.6 & \underline{50.9} & \underline{45.7} \\
\midrule
DMKGC & \bf 41.9 & \bf 91.0 & \bf 61.0 & \bf 20.9 & \bf 64.2 & \bf 36.3 & \bf 33.0 & \bf 77.7 & \bf 49.1 & \bf 38.6 & \bf 82.3 & \bf 54.2 & \bf 42.0 & \bf 82.4 & \bf 56.3 & \bf 51.4 \\
\bottomrule
\end{tabular*}
\label{Tab:MainDBP5L}
\end{table*}

\begin{table*}[!t]
\centering
\small
\caption{Results (\%) on E-PKG. $\dag$ indicates re-produced results. The best result is \textbf{bold-faced} and the runner-up is \underline{underlined}. }
\setlength{\tabcolsep}{3pt}
\begin{tabular*}{1 \textwidth}{@{\extracolsep{\fill}}@{}lccc|ccc|ccc|ccc|ccc|ccc|c@{}}
\toprule
\multirow{2.5}{*}{\bf Method} & \multicolumn{3}{c}{\bf DE} & \multicolumn{3}{c}{\bf EN} &  \multicolumn{3}{c}{\bf ES} & \multicolumn{3}{c}{\bf FR} & \multicolumn{3}{c}{\bf IT} &  \multicolumn{3}{c}{\bf JA}&{\bf AVG}  \\
\cmidrule{2-4}\cmidrule{5-7}\cmidrule{8-10} \cmidrule{11-13}\cmidrule{14-16}\cmidrule{17-19}\cmidrule{20-20}
& H@1  & H@10   &  MRR  
& H@1  & H@10   &  MRR  
& H@1  & H@10   &  MRR     
& H@1  & H@10   &  MRR     
& H@1  & H@10   &  MRR     
& H@1  & H@10   &  MRR   & MRR
\\
\midrule
TransE  & 21.2 & 65.5 & 37.4 & 23.2 & 67.5 & 39.4 & 17.2 & 58.4 & 33.0 & 20.8 & 66.9 & 37.5 & 22.0 & 63.8 & 37.8 & 25.1 & 72.7 & 43.6    & 38.1
\\
DistMult & 21.4 & 54.5 & 35.4 & 23.8 & 60.1 & 37.2 & 17.9 & 46.2 & 30.9 & 20.7 & 53.5 & 35.1 & 22.8 & 51.8 & 34.8 & 25.9 & 62.6 & 38.0   & 35.2
\\
RotatE & 22.3 & 64.3 & 38.2 & 24.2 & 66.8 & 40.0 & 18.3 & 58.9 & 33.7 & 22.1 & 64.3 & 38.2 & 22.5 & 64.0 & 38.1 & 26.3 & 71.9 & 41.8   & 38.3
\\
KG-BERT & 21.8 & 64.7 & 38.4 & 24.3 & 66.4 & 39.6 & 18.7 & 58.8 & 33.2 & 22.3 & 67.2 & 38.3 & 22.9 & 63.7 & 37.2 & 26.9 & 72.4 & 44.1    & 38.5\\
\midrule
KEnS & 24.3 & 65.8 & - & 26.2 & 69.5 & - & 21.3 & 59.5 & - & 25.4 & 68.2 & - & 25.1 & \underline{64.6} & - & 33.5 & 73.6 & -   & - \\
CG-MuA & 22.9 & 64.9 & 38.7 & 24.8 & 67.9 & 40.2 & 19.2 & 58.8 & 33.8 & 23.0 & 67.5 & 39.1 & 23.9 & 63.8 & 37.6 & 30.4 & 72.9 & 45.9  & 39.2\\
AlignKGC & 22.1 & 65.1 & 38.5 & 25.6 & 68.3 & 40.5 & 19.4 & 59.1 & 34.2 & 22.8 & 67.2 & 38.8 & 24.2 & 63.4 & 37.3 & 31.2 & 72.3 & 46.2    & 39.3
\\
SS-AGA & 24.6 & 66.3 & 39.4 & 26.7 & 69.8 & 41.5 & 21.0 & 60.1 & 36.3 & \underline{25.9} & \underline{68.7} & \underline{40.2} & 24.9 & 63.8 & 38.4 & 33.9 & 74.1 & 48.3    & 40.7
\\
LSMGA & \underline{30.7} & \underline{68.5} & \underline{44.8} & \underline{31.9} & \underline{70.2} & \underline{45.9} & {23.1} & \underline{61.1} & {36.5} & {23.7} & 63.5 & 38.2 & {26.8} & {64.5} & \underline{41.0} & {43.7} & \underline{78.4} & {57.1}  & \underline{43.9} \\
GLKGC$^\dag$ & 24.1 & 63.6 & 37.7 & 27.1 & 58.4 & 39.4 & \underline{24.6} & 61.0 & \underline{36.8} & 22.1 & 62.3 & 36.4 & \underline{27.0} & 63.7 & 40.4 & \underline{44.1} & 76.4 & \underline{57.5} & 41.4 \\
\midrule
DMKGC & \bf 30.9 & \bf 69.1 & \bf 45.0 & \bf 33.3 & \bf 70.3 & \bf 46.8 & \bf 26.7 & \bf 63.7 & \bf 39.7 & \bf 26.0 & \bf 68.8 & \bf 40.7 & \bf 31.2 & \bf 66.3 & \bf 44.6 & \bf 50.1 & \bf 79.1 & \bf 61.8 & \bf 46.4 \\
\bottomrule
\end{tabular*}
\label{Tab:MainEPKG}
\end{table*}
\begin{table}[!t]
\small
\centering
\caption{Results (\%) on DWY. $\dag$ indicates re-produced results. The best result is \textbf{bold-faced} and the runner-up is \underline{underlined}.}
\setlength{\tabcolsep}{1pt}
\begin{tabular*}{0.48\textwidth}{@{\extracolsep{\fill}}@{}lccc|ccc|ccc|c@{}}
\toprule
\multirow{2.5}{*}{\bf Method} & \multicolumn{3}{c}{\bf DB} & \multicolumn{3}{c}{\bf WK} &  \multicolumn{3}{c}{\bf YG}&\bf AVG   \\
\cmidrule{2-4}\cmidrule{5-7}\cmidrule{8-10}\cmidrule{11-11}
& H@1  & H@10   &  MRR  
& H@1  & H@10   &  MRR  
& H@1  & H@10   &  MRR  & MRR
 \\
\midrule
TransE$^\dag$  & 4.3  & 52.9 & 20.3 &  3.0 & 48.6 & 17.3 & 2.2 & 42.2 & 13.1 & 16.9 \\
DistMult$^\dag$  & 8.6 & 36.5 & 17.6 & 8.4 & 41.7 & 18.4 & 4.6 & 32.5  & 12.7 & 16.2 \\
RotatE$^\dag$  & 13.2  & 57.4 & 27.9 & \underline{9.9} & 52.5 & 26.4 & 3.5  & 42.7  & 13.8  & 22.7 \\
\midrule
SS-AGA$^\dag$  & 5.8 & 61.8 & 22.6 & 6.6 & 52.2 & 18.5  & 9.0  & 52.3 & 22.9  & 21.3 \\
LSGMA$^\dag$ & \underline{14.0}  & 64.3  &  30.9 & {9.5} & 54.6 & 23.9 & 11.4 & 48.6 &  23.5  & 26.1 \\
GLKGC$^\dag$ & 13.4 & \underline{66.9} & \underline{32.3} & 9.3 & \underline{55.0} & \underline{24.3} & \underline{16.5} &  \underline{52.8} & \underline{28.7} & \underline{28.4} \\
\midrule
DMKGC & \bf 15.7 & \bf 68.9 & \bf 34.5 & \bf 11.4 & \bf 59.3 & \bf 26.7 & \bf 23.6 & \bf 66.7 & \bf 37.9 & \bf 33.1 \\

\bottomrule
\end{tabular*}

\label{Tab:MainDWY}
\end{table}

\subsubsection{Datasets}
For evaluation, we use three benchmarks comprising 14 KGs: the multilingual \textbf{DBP-5L}~\cite{Chen2017:MTransE} and \textbf{E-PKG}~\cite{Huang2022:SS-AGA}, and a constructed multi-domain \textbf{DWY}~\cite{BootEA} dataset.
DBP-5L contains five DBpedia-based KGs in Greek (EL), English (EN), Spanish (ES), French (FR), and Japanese (JA).
Besides, E-PKG includes industrial e-commerce mobile phone data in six languages: German (DE), English (EN), Spanish (ES), French (FR), Italian (IT), and Japanese (JA).
In addition, DWY integrates DBpedia (DB), YAGO (YG), and Wiki (WK), where we adopt the original aligned entities between each two KGs~\cite{BootEA}, and take 80\%, 10\%, 10\% triples in each KG for training, validation and testing.
All datasets provide aligned entity pairs between KGs, with unified and shared relations across all KGs~\cite{Chen2017:MTransE, Tang2023:LSMGA}.
The statistics are shown in \textbf{Appendix}.

\subsubsection{Baselines}

To evaluate our model, we select the following state-of-the-art methods as baselines:
(i) \textbf{Single-domain methods}, which perform inference within individual KGs without knowledge transfer between KGs, including \textbf{TransE}~\cite{transE}, \textbf{DisMult}~\cite{dismult}, \textbf{RotatE}~\cite{rotatE}, \textbf{KG-BERT}~\cite{KGbert}.
(ii) \textbf{Multi-domain methods}, which mostly attempt consistency-based modules to transfer knowledge from support KGs for target KG predictions, including \textbf{KEnS}~\cite{KEnS:chen-etal-2020-multilingual}, \textbf{CG-MuA}~\cite{CG_MuAlign}, \textbf{AlignKGC}~\cite{Singh2021:AlignKGC}, \textbf{SS-AGA}~\cite{Huang2022:SS-AGA}, \textbf{LSMGA}~\cite{Tang2023:LSMGA} and \textbf{GLKGC}~\cite{GLMKGC}.
For details, please refer to \textbf{Appendix}.

\subsubsection{Evaluation Protocol}
Following previous studies~\cite{Chen2017:MTransE,Tang2023:LSMGA}, we evaluate models in the task of \textit{tail entity prediction}.
During training, we combine all the training data from the multiple KGs.
In testing, we rank all candidate entities of the target KG to predict $t$ given $h$ and $r$ for each triple $(h,r,?)$ in the test data.
Three metrics are reported, including Hits@10 (H@10 for short), Hits@1 (H@1) and mean reciprocal ranks (MRR).
Following~\citet{Tang2023:LSMGA}, the optimal model is selected according to the average MRR of all KGs.

\subsubsection{Implementation Details}
Most hyperparameters are shared for all datasets. 
The entity and relation embeddings are randomly initialized with dimension 256.
The learning rate is set to 0.001, and the margin $\lambda$ is set to 0.5 for all datasets.
The KG encoder has 2 layers.
The diffusion step $T$ is selected in $\{2,4,\cdots,64\}$, and the strength $s$ in $\{1,2,\cdots,5\}$, $\omega_1,\omega_2$ are tuned in $\{1,3,5\}\!\times\!10^{-\{4,3,2,1\}}$.
For baselines, most results on DBP-5L and E-PKG are obtained from original literature.
On DWY, we re-implement baselines with the best hyperparameters reported.
We employ a grid search with three trials, and the optimal hyperparameters are reported in \textbf{Appendix}.

\subsection{Main Results (RQ1)}
\subsubsection{Method Comparison}
We present the comparison between our model and existing baselines in Table~\ref{Tab:MainDBP5L}, \ref{Tab:MainEPKG} and \ref{Tab:MainDWY}.
We find that:
\textit{\underline{First}, multi-domain KGC methods outperform single-domain methods.}
This ensures the effectiveness of using support KGs for target KG prediction, which can help improve inference on entities with limited triples.
\textit{\underline{Second}, our model outperforms all existing methods.}
Specifically, our model achieves average improvements in MRR of 5.7\%, 2.5\%, and 4.7\% on DBP-5L, E-PKG, and DWY, respectively.
This reflects the effectiveness of using conditional diffusion to achieve knowledge transfer, which produces more informative embeddings than existing multi-domain KGC methods, such as SS-AGA, LSGMA and GLKGC. 
\textit{\underline{Third}, our model obtains consistent improvements in three datasets with 14 KGs, with an overall 4.3\% MRR improvement}.
The results on multilingual DBP-5L, industrial E-PKG, and multi-domain DWY broadly demonstrate the generality of our model.

\begin{table}[!tbp]
\small
\caption{Variant analysis on DBP-5L, where AVG-H@1, AVG-H@10 and AVG-MRR denote the average metrics (\%).}
\centering\setlength{\tabcolsep}{3pt}
\begin{tabular*}{0.98\columnwidth}{@{\extracolsep{\fill}}@{}l|cccc@{}}
\toprule
\bf Variant & \bf AVG-H@1 & \bf AVG-H@10 & \bf AVG-MRR & \bf $\Delta$AVG-MRR  \\ 
\midrule
Entire & \bf 35.3 &	\bf 79.5 & \bf 51.4 & - \\
\midrule
repl. GCN & 23.2 & 69.8 & 39.7 & $\downarrow$ 11.7 \\
\midrule
w/o cond & 27.0 & 76.3 & 44.9 & $\downarrow$ 6.5 \\
repl. mean & 32.6 & 71.6 & 46.4 & $\downarrow$ 5.0 \\
\midrule
w/o DM & 31.6 & 68.9 & 44.4 & $\downarrow$ 7.0 \\
w/o reg & 32.9 & 77.8 & 49.2 & $\downarrow$ 2.2 \\
\midrule
repl. $\epsilon$-ELBO & 18.8 & 70.8 & 37.3 & $\downarrow$ 14.1 \\
repl. $\epsilon$-init & 28.1 & 77.2 & 45.9 &  $\downarrow$ 5.5 \\
repl. cosine & 25.7 & 74.2 & 43.0 & $\downarrow$ 8.4 \\
\bottomrule
\end{tabular*}
\label{Tab:VariantAnalysis}
\end{table}

\begin{table}[!tbp]
\small
\caption{Comparison with general knowledge transfer methods. The results (\%) are reported on DBP-5L.}
\centering\setlength{\tabcolsep}{3pt}
\begin{tabular*}{0.98\columnwidth}{@{\extracolsep{\fill}}@{}l|cccc@{}}
\toprule
\bf Method & \bf AVG-H@1 & \bf AVG-H@10 & \bf AVG-MRR   \\ 
\midrule
InfoNCE$^\dag$ & 31.5 & 71.4 & 45.3 \\
DA-DIFF$^\dag$ & 32.6 & 71.4 & 46.3  \\
MMD$^\dag$ &  34.6 & 74.3 & 48.4 \\
\midrule
DMKGC & \bf 35.3 &	\bf 79.5 & \bf 51.4  \\
\bottomrule
\end{tabular*}
\label{Tab:AnalysisKT}
\end{table}

\subsubsection{In-depth Variant Analysis}
To evaluate the unity of the components, we perform variant analysis in Table~\ref{Tab:VariantAnalysis}.
We find that:
\textit{\underline{First}, repl. GCN replaces the attentive KG encoder with non-relational GCN~\cite{GCN}}, indicating the fundamental role of a capable encoder.
\textit{\underline{Second}, w/o cond removes the condition in diffusion, and repl. mean replaces the attentive condition fuser with simple vector mean}.
The results reflect the crucial unity of the condition information.
\textit{\underline{Third}, w/o DM removes the diffusion knowledge transfer and w/o reg removes the regularization $\mathcal{L}_{\mathrm{reg}}$}.
This reflects the vital role of learning domain-general entity embeddings with unbiased generation.
\textit{\underline{Forth}, repl. $\epsilon$-ELBO achieves ELBO with DDPM~\cite{DDPM:NIPS2020}, repl. $\epsilon$-init starts the reverse process with a random noise, and repl. cosine replaces L2-norm with cosine score in Eq.~(\ref{Eq:L_unified}).}
The results reflect that direct parameterizing $\bm{x}_0$ by $\hat{\bm{x}}_\theta$ can be helpful in learning the general entity embedding, the initial $\bm{x}_T$ achieves a better initialization, and cosine depicts relative embedding similarity but can hardly reconstruct the embedding details.
\textit{All results indicate the effectiveness of components}.

\subsubsection{Analysis on Knowledge Transfer}
To evaluate the impact of conditional diffusion transfer, we adapt three general knowledge transfer methods to our KG encoder backbone, and fuse the refined embeddings with mean fusion:
(i) \textbf{InfoNCE}~\cite{SimCLR:2020} leverages contrastive loss to enhance the consistency between equivalent entities from various KGs.
(ii) \textbf{DA-DIFF}~\cite{DA-DIFF:TIP24} uses the fused support KG embedding as the initial embedding, adds noise to it, and learns to approximate the also noised target KG embedding step by step.
(iii) \textbf{MMD}~\cite{MMD:2012} adopts the classical maximum mean discrepancy, which measures the distributional relevance between equivalent entities.
The results are shown in Table~\ref{Tab:AnalysisKT}.
We find that our model outperforms the three methods.
Actually, the three methods commonly capture the consistency between equivalent entities in different aspects.
We believe that \textit{our conditional diffusion transfer learns to generate a domain-general entity embedding, thus allowing more informative embeddings for task prediction in various KGs.}

\subsection{Results in Low-resource Scenarios (RQ2)}\label{Sec:Exp_Low_Res}

\begin{figure}[!t]
	\begin{center}
            \includegraphics[width=0.98\columnwidth]{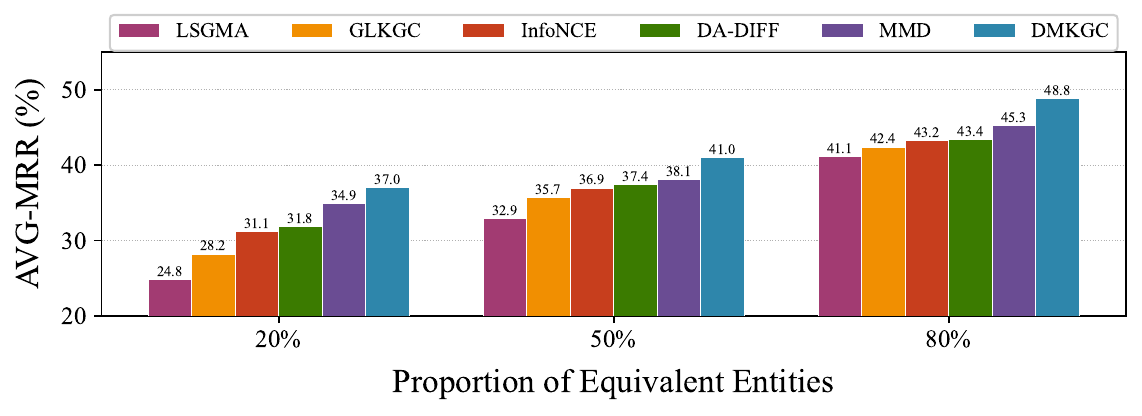}
		\caption{Results (\%) with a limited number of equivalent entities, randomly selected at 20\%, 50\% and 80\% on DBP-5L.}
		\label{Exp:Seed}
	\end{center}
\end{figure}

\begin{figure}[!t]
	\begin{center}
            \includegraphics[width=0.98\columnwidth]{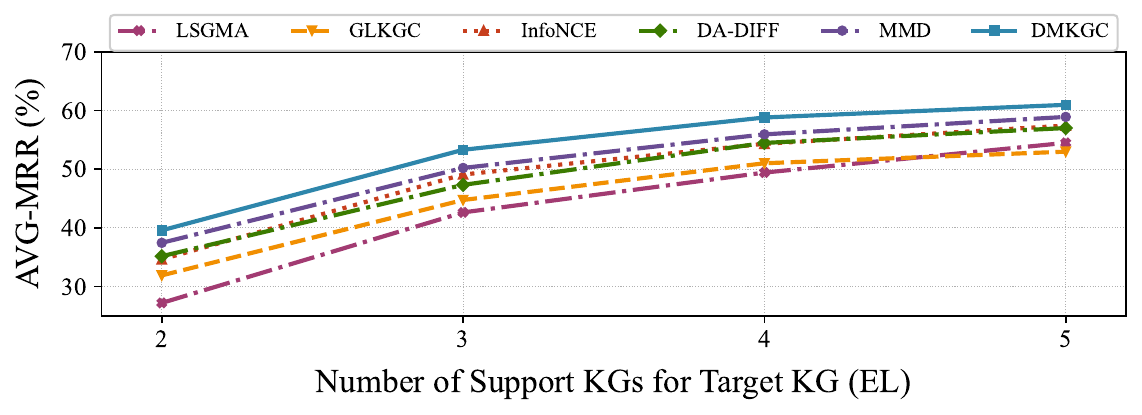}
		\caption{Results (\%) with a limited number of support KGs for target KG prediction (EL) on DBP-5L.}
		\label{Exp:nKG}
	\end{center}
\end{figure}

\begin{table*}[!t]
\small 
\caption{Results (\%) on unseen entity settings conducted on DBP-5L, where head entities in queries are unseen in training. }
\centering
\setlength{\tabcolsep}{3pt}
\begin{tabular*}{1 \textwidth}{@{\extracolsep{\fill}}@{}lccc|ccc|ccc|ccc|ccc|c@{}}
\toprule
\multirow{2.5}{*}{\bf Methods} & \multicolumn{3}{c}{\bf EL} & \multicolumn{3}{c}{\bf EN} &  \multicolumn{3}{c}{\bf ES} & \multicolumn{3}{c}{\bf FR} & \multicolumn{3}{c}{\bf JA} &\bf AVG \\
\cmidrule{2-4}\cmidrule{5-7}\cmidrule{8-10} \cmidrule{11-13}\cmidrule{14-16}\cmidrule{17-17}
& H@1  & H@10   &  MRR  
& H@1  & H@10   &  MRR  
& H@1  & H@10   &  MRR     
& H@1  & H@10   &  MRR     
& H@1  & H@10   &  MRR & MRR
 \\
\midrule
LSGMA$^\dag$ & 7.1 & 48.8 & 21.7 & 4.5 & 26.0 & 12.0 & 7.5 & 37.5 & 17.6 & 7.8 & 42.3 & 20.2 & 7.0 & 33.8 & 16.7 & 17.7 \\
GLKGC$^\dag$ & 15.9 & 57.6 & 31.5 & 5.3 & 29.9 & 14.0 & 9.7 & 42.2 & 21.3 & 13.6 & 47.3 & 26.0 & 10.0 & 40.1 & 21.2 & 22.8  \\
InfoNCE$^\dag$ & 21.6 & \underline{57.8} & \underline{34.0} & 9.3 & 28.6 & 15.8 & 13.4 & 39.5 & 22.4 & 19.8 & 46.3 & 29.2 & 17.2 & 39.5 & 25.1 & 25.3  \\
DA-DIFF$^\dag$ & 12.2 & 54.8 & 27.5 & 5.9 & \underline{31.3} & 14.8 & 11.0 & \underline{45.7} & \underline{22.6} & 15.9 & \underline{50.2} & 29.5 & 9.4 & \underline{43.0} & 22.0 & 23.3  \\
MMD$^\dag$ & \underline{22.1} & 55.3 & 33.9 & \underline{9.6} & 28.1 & \underline{16.1} & \underline{13.8} & 39.1 & 22.5 & \underline{20.7} & 46.1 & \underline{30.0} & \underline{17.3} & 39.3 & \underline{25.2} & \underline{25.5} \\
\midrule
DMKGC & \bf 24.9 & \bf 61.8 & \bf \bf 39.0 & \bf 12.4 & \bf 32.1 & \bf 19.4 & \bf 19.0 & \bf 46.2 & \bf 28.8 & \bf 23.4 & \bf 50.4 & \bf 33.4 & \bf 19.3 & \bf 43.3 & \bf 27.6 & \bf 29.6 \\
\bottomrule
\end{tabular*}
\label{Tab:Unseen}
\end{table*}

\subsubsection{Analysis on Limited Equivalent Entities}
To investigate robustness, we performed experiments with limited equivalent entities (20\%, 50\%, 80\%), shown in Figure~\ref{Exp:Seed}.
We find that \textit{our model consistently outperforms existing methods.}
With fewer equivalent entities, the methods have less supervision to learn knowledge transfer.
In contrast to consistency-based methods that rely on the given equivalent entities, \textit{our model uses prior entity embeddings as the supervision, which learns domain-general information for generation using all entities in domains, enriching knowledge transfer.}

\subsubsection{Analysis on Fewer Support KGs}
For investigation, we also explored experiments with various numbers of support KGs, shown in Figure~\ref{Exp:nKG}.
We find that:
\textit{\underline{First}, more support KGs lead to better results in the target KG.}
This indicates the significance of the MKGC task in improving low-resource KGC.
\textit{\underline{Second}, our model achieves better results for all numbers.}
We believe that our method learns domain-general entity information, which produces more informative embeddings than the existing consistency-based methods.

\subsubsection{Analysis on Unseen Entities}
To further investigate generalizability, we conducted experiments on unseen entities in an extrapolation setting~\cite{KnowledgeExtrapolationSurvey:IJCAI23}.
Specifically,\textit{ for a query $( h, r, ?)$, we suppose that $h$ is unseen in the target KG, but it has triples in other support KGs.}
This setting is useful for new-coming entities to build connections in the current KG.
To implement it, for head entities in the testing data, we remove their triples in the training data of the located KG, thus, the prediction has to rely on the related triples from other KGs.
The results are shown in Table~\ref{Tab:Unseen}.
We find that our model derives the best results and significantly exceeds the previous baselines LSGMA and GLKGC.
We believe that\textit{ the diffusion generation module learns general information about entities, which is representative for unseen entities.}
We leave further studies in future works.

\section{Related Works}

\subsection{Knowledge Graph Completion}
\textit{Knowledge graph completion (KGC)} aims to predict missing triples based on existing triples in a single KG. 
Classical studies propose \textit{triple-based methods}~\cite{transE,dismult,rotatE,ConvE,complEx} with translation-based~\cite{transE,rotatE} or semantic matching-based score functions~\cite{dismult, ConvE}. 
Later studies propose \textit{GNN-based methods}~\cite{sacn,R-GCN, KNOWFORMER} to capture relational graph structures.
Recent studies also explore entity textual information~\cite{DKRL:AAAI16,DBLP:conf/icassp/YaoPM025} with PLMs~\cite{KGbert,KEPLER:TACL21} or LLMs~\cite{KICGPT:emnlp2023,MKGL:NeurIPS24,KoPA:MM24,DIFT:ISWC24,DBLP:conf/icassp/YaoPM025,CDKGC:EACL24,DBLP:conf/icassp/YaoPM025}, which are not designed in our task setting.
Overall, these studies attempt to achieve KGC in an individual KG, which cannot be directly used for multi-domain and low-resource scenarios.

\subsection{Multi-domain KG Completion}
\textit{Multi-domain KG completion (MKGC)} aims to fully utilize multiple KG triples to improve KGC.
Early studies explore it exclusively in multilingual scenarios, namely \textit{multilingual KGC}~\cite{Huang2022:SS-AGA,Tang2023:LSMGA}.
For generality, we term it as multi-domain KGC beyond languages.
For the methods, MTransE~\cite{Chen2017:MTransE} first extends the KG embeddings from one to multiple KGs.
Later studies~\cite{MultiViewEA:IJCAI19, RNeiEA:AAAI21,ICML23:multisource, yang2025translationMEA} focus mainly on entity alignment (EA) for knowledge fusion.
Further studies~\cite{CG_MuAlign,Singh2021:AlignKGC,Huang2022:SS-AGA,GLMKGC,sheng2026:IMKGC} explore KGC with the other related KGs. 
They encode KGs with relational GNNs, and leverage EA for consistency to transfer knowledge for the target KGC.
However, consistency-based methods can limit entity representation in knowledge transfer, impeding domain-specific information.
To our knowledge, few studies have explored generation-based transfer for MKGC.

\subsection{Diffusion Models}
\textit{Diffusion models (DMs)} have achieved remarkable success in generative tasks such as image~\cite{SDXL:ICLR2024,DDPM:NIPS2020} and text generation~\cite{DiffLang:NIPS23,LDLang:NIPS23}, and also reflect potential in discriminative tasks~\cite{Seg:CVPR23,DiffusionRM,DreamRec:NeurIPS2023,DiffKG:WSDM24}.
Recent KGC studies attempt DMs to model the generative distribution of triples~\cite{FactEmbedding:WWW24, KGDM:AAAI24, DiffusionCom:25} and graph structures~\cite{DiffusionE:KDD24} for triple prediction.
Our paper explores \textit{knowledge transfer for multi-domain KGC}, which has a different research focus.
For knowledge transfer, existing studies explore DMs for knowledge distillation~\cite{DiffKD:NIPS23,DiffDomainTea} or domain adaptation~\cite{DA-DIFF:TIP24}, which transfer knowledge mainly through a denoising process using the diffusion path as a bridge. 
Unlike them, our paper leverages diffusion models to pioneer the unbiased generation from support domains and simultaneously allows target information for prediction, making a more informative knowledge transfer.

\section{Conclusion}

This paper addresses MKGC, which transfers knowledge from support KGs to improve KGC in a target KG.
Existing studies mainly leverage consistency-based methods, potentially surpassing domain-specific KG information.
To address this, we propose a novel generation-based paradigm.
Our framework, DMKGC, uses conditional diffusion models to generate domain-general entity embeddings, effectively integrating support KG knowledge while preserving domain-specific information. 
By treating each KG as a partial view of entities and using domain-agnostic embeddings as unbiased generation targets, DMKGC learns rich generalizable representations.
Experiments on 14 KGs show that DMKGC achieves significant gains and consistently excels in low-resource scenarios.

\section*{Acknowledgments}
The authors thank the reviewers for their helpful feedback.
This work was supported by the National Natural Science Foundation of China (No. 62406319).

\bibliographystyle{ACM-Reference-Format}
\bibliography{sample-base}

@String{Computing = "Computing" }

@String{Computer = "{IEEE} Computer" }

@String{Springer = "Springer-Verlag" }

@ArtifactSoftware{R,
    title = {R: A Language and Environment for Statistical Computing},
    author = {{R Core Team}},
    organization = {R Foundation for Statistical Computing},
    address = {Vienna, Austria},
    year = {2019},
    url = {https://www.R-project.org/},
}

@inproceedings{Chen2020:KEnS,
  title={Multilingual Knowledge Graph Completion via Ensemble Knowledge Transfer},
  author={Chen, Xuelu and Chen, Muhao and Fan, Changjun and Uppunda, Ankith and Sun, Yizhou and Zaniolo, Carlo},
  booktitle={Findings of EMNLP},
  pages={3227--3238},
  year={2020}
}

@inproceedings{CG_MuAlign,
author = {Zhu, Qi and Wei, Hao and Sisman, Bunyamin and Zheng, Da and Faloutsos, Christos and Dong, Xin Luna and Han, Jiawei},
title = {Collective Multi-Type Entity Alignment Between Knowledge Graphs},
year = {2020},
booktitle = {Proceedings of Web Conference}
}

@inproceedings{Chen2017:MTransE,
    title={Multilingual Knowledge Graph Embeddings for Cross-lingual Knowledge Alignment},
    author={Chen, Muhao and Tian, Yingtao and Yang, Mohan and Zaniolo, Carlo},
    booktitle={Proceedings of IJCAI},
    year={2017}
}

@article{Singh2021:AlignKGC,
  title={Multilingual Knowledge Graph Completion with Joint Relation and Entity Alignment},
  author={Singh, Harkanwar and Jain, Prachi and Chakrabarti, Soumen and others},
  journal={arXiv preprint arXiv:2104.08804},
  year={2021}
}

@inproceedings{transE,
author = {Bordes, Antoine and Usunier, Nicolas and Garcia-Dur\'{a}n, Alberto and Weston, Jason and Yakhnenko, Oksana},
title = {Translating Embeddings for Modeling Multi-Relational Data},
year = {2013},
booktitle = {Proceedings of NeurIPS},
pages = {2787–2795},
numpages = {9},
}

@inproceedings{rotatE,
 title={RotatE: Knowledge Graph Embedding by Relational Rotation in Complex Space},
 author={Zhiqing Sun and Zhi-Hong Deng and Jian-Yun Nie and Jian Tang},
 booktitle={Proceedings of ICLR},
 year={2019}
}

@article{KGbert,
  title={KG-BERT: BERT for Knowledge Graph Completion},
  author={Liang Yao and Chengsheng Mao and Yuan Luo},
  journal={Proceedings of AAAI},
  year={2020},
}

@inproceedings{ConvE,
	author = {Dettmers, Tim and Pasquale, Minervini and Pontus, Stenetorp and Riedel, Sebastian},
	Booktitle = {Proceedings of AAAI},
	Title = {Convolutional 2D Knowledge Graph Embeddings},
	Year = {2018},
    pages  = {1811--1818}
}

@inproceedings{dismult,
author = {Yang, Bishan and Yih, Wen-tau and He, Xiaodong and Gao, Jianfeng and Deng, Li},
title = {Embedding entities and
relations for learning and inference in knowledge
bases},
year = {2015},
booktitle = {Proceedings of ICLR},
}

@inproceedings{GCN,
  title={Semi-Supervised Classification with Graph Convolutional Networks},
  author={Kipf, Thomas N. and Welling, Max},
  booktitle={International Conference on Learning Representations},
  year={2017}
}

@inproceedings{adam,
title={Adam: A
method for stochastic optimization.},
author={Diederik P Kingma and Jimmy Ba},
booktitle={International Conference on Learning Representations},
year={2014},
}

@inproceedings{Huang2022:SS-AGA,
  author       = {Zijie Huang and
                  Zheng Li and
                  Haoming Jiang and
                  Tianyu Cao and
                  Hanqing Lu and
                  Bing Yin and
                  Karthik Subbian and
                  Yizhou Sun and
                  Wei Wang},
  title        = {Multilingual Knowledge Graph Completion with Self-Supervised Adaptive
                  Graph Alignment},
  booktitle    = {Proceedings of ACL},
  pages        = {474--485},
  year         = {2022},
  url          = {https://doi.org/10.18653/v1/2022.acl-long.36},
  doi          = {10.18653/V1/2022.ACL-LONG.36},
  bibsource    = {dblp computer science bibliography, https://dblp.org}
}

@inproceedings{Tang2023:LSMGA,
  author       = {Rongchuan Tang and
                  Yang Zhao and
                  Chengqing Zong and
                  Yu Zhou},
  title        = {Multilingual Knowledge Graph Completion with Language-Sensitive Multi-Graph
                  Attention},
  booktitle    = {Proceedings of ACL},
  pages        = {10508--10519},
  year         = {2023},
  url          = {https://doi.org/10.18653/v1/2023.acl-long.586},
  doi          = {10.18653/V1/2023.ACL-LONG.586},
  bibsource    = {dblp computer science bibliography, https://dblp.org}
}

@inproceedings{vae,
  title={Auto-Encoding Variational Bayes},
  author={Kingma, Diederik P and Welling, Max},
  booktitle={Proceedings of ICLR},
  year={2014}
}

@ARTICLE{tkde_survey_llmkg_pan,
  author={Pan, Shirui and Luo, Linhao and Wang, Yufei and Chen, Chen and Wang, Jiapu and Wu, Xindong},
  journal={IEEE Transactions on Knowledge and Data Engineering}, 
  title={Unifying Large Language Models and Knowledge Graphs: A Roadmap}, 
  year={2024},
  volume={36},
  number={7},
  pages={3580-3599},
  doi={10.1109/TKDE.2024.3352100}}

@ARTICLE{wang2017:kge,
  author={Wang, Quan and Mao, Zhendong and Wang, Bin and Guo, Li},
  journal={IEEE Transactions on Knowledge and Data Engineering}, 
  title={Knowledge Graph Embedding: A Survey of Approaches and Applications}, 
  year={2017},
  volume={29},
  number={12},
  pages={2724-2743},
  doi={10.1109/TKDE.2017.2754499}}

@inproceedings{sacn,
  title={End-to-end structure-aware convolutional networks for knowledge base completion},
  author={Shang, Chao and Tang, Yun and Huang, Jing and Bi, Jinbo and He, Xiaodong and Zhou, Bowen},
  booktitle={Proceedings of AAAI},
  volume={33},
  number={01},
  pages={3060--3067},
  year={2019}
}

@inproceedings{KNOWFORMER,
author = {Liu, Junnan and Mao, Qianren and Jiang, Weifeng and Li, Jianxin},
title = {KNOWFORMER: revisiting transformers for knowledge graph reasoning},
year = {2024},
booktitle = {Proceedings of ICML},
articleno = {1280},
numpages = {22},
location = {Vienna, Austria},
series = {ICML'24}
}

@inproceedings{R-GCN,
  title={Modeling relational data with graph convolutional networks},
  author={Schlichtkrull, Michael and Kipf, Thomas N and Bloem, Peter and Van Den Berg, Rianne and Titov, Ivan and Welling, Max},
  booktitle={European semantic web conference},
  pages={593--607},
  year={2018},
  organization={Springer}
}

@inproceedings{KICGPT:emnlp2023,
  author       = {Yanbin Wei and
                  Qiushi Huang and
                  Yu Zhang and
                  James T. Kwok},
  title        = {{KICGPT:} Large Language Model with Knowledge in Context for Knowledge
                  Graph Completion},
  booktitle    = {Findings of {EMNLP}},
  pages        = {8667--8683},
  year         = {2023},
  url          = {https://doi.org/10.18653/v1/2023.findings-emnlp.580},
  doi          = {10.18653/V1/2023.FINDINGS-EMNLP.580},
  bibsource    = {dblp computer science bibliography, https://dblp.org}
}

@inproceedings{CDKGC:EACL24,
  author       = {Dawei Li and
                  Zhen Tan and
                  Tianlong Chen and
                  Huan Liu},
  editor       = {Yvette Graham and
                  Matthew Purver},
  title        = {Contextualization Distillation from Large Language Model for Knowledge
                  Graph Completion},
  pages        = {458--477},
  year         = {2024},
  url          = {https://aclanthology.org/2024.findings-eacl.32},
  bibsource    = {dblp computer science bibliography, https://dblp.org}
}

@inproceedings{DBLP:conf/icassp/YaoPM025,
  author       = {Liang Yao and
                  Jiazhen Peng and
                  Chengsheng Mao and
                  Yuan Luo},
  title        = {Exploring Large Language Models for Knowledge Graph Completion},
  booktitle    = {Proceedings of {IEEE} {ICASSP}},
  pages        = {1--5},
  publisher    = {{IEEE}},
  year         = {2025},
  url          = {https://doi.org/10.1109/ICASSP49660.2025.10889242},
  doi          = {10.1109/ICASSP49660.2025.10889242},
  bibsource    = {dblp computer science bibliography, https://dblp.org}
}

@inproceedings{DIFT:ISWC24,
  author       = {Yang Liu and
                  Xiaobin Tian and
                  Zequn Sun and
                  Wei Hu},
  title        = {Finetuning Generative Large Language Models with Discrimination Instructions for Knowledge Graph Completion},
  booktitle    = {Proceedings of ISWC},
  pages        = {199--217},
  year         = {2024},
}

@article{KEPLER:TACL21,
  author       = {Xiaozhi Wang and
                  Tianyu Gao and
                  Zhaocheng Zhu and
                  Zhengyan Zhang and
                  Zhiyuan Liu and
                  Juanzi Li and
                  Jian Tang},
  title        = {{KEPLER:} {A} Unified Model for Knowledge Embedding and Pre-trained
                  Language Representation},
  journal      = {TACL},
  volume       = {9},
  pages        = {176--194},
  year         = {2021},
  url          = {https://doi.org/10.1162/tacl\_a\_00360},
  doi          = {10.1162/TACL\_A\_00360},
  bibsource    = {dblp computer science bibliography, https://dblp.org}
}

@inproceedings{DKRL:AAAI16,
  author       = {Ruobing Xie and
                  Zhiyuan Liu and
                  Jia Jia and
                  Huanbo Luan and
                  Maosong Sun},
  editor       = {Dale Schuurmans and
                  Michael P. Wellman},
  title        = {Representation Learning of Knowledge Graphs with Entity Descriptions},
  booktitle    = {Proceedings of AAAI},
  pages        = {2659--2665},
  year         = {2016},
  url          = {https://doi.org/10.1609/aaai.v30i1.10329},
  doi          = {10.1609/AAAI.V30I1.10329},
  bibsource    = {dblp computer science bibliography, https://dblp.org}
}

@inproceedings{MultiViewEA:IJCAI19,
  author       = {Qingheng Zhang and
                  Zequn Sun and
                  Wei Hu and
                  Muhao Chen and
                  Lingbing Guo and
                  Yuzhong Qu},
  editor       = {Sarit Kraus},
  title        = {Multi-view Knowledge Graph Embedding for Entity Alignment},
  booktitle    = {Proceedings of IJCAI},
  pages        = {5429--5435},
  year         = {2019},
  url          = {https://doi.org/10.24963/ijcai.2019/754},
  doi          = {10.24963/IJCAI.2019/754},
  bibsource    = {dblp computer science bibliography, https://dblp.org}
}

@inproceedings{RNeiEA:AAAI21,
  author       = {Yao Zhu and
                  Hongzhi Liu and
                  Zhonghai Wu and
                  Yingpeng Du},
  title        = {Relation-Aware Neighborhood Matching Model for Entity Alignment},
  booktitle    = {Proceedings of AAAI},
  pages        = {4749--4756},
  year         = {2021},
  url          = {https://doi.org/10.1609/aaai.v35i5.16606},
  doi          = {10.1609/AAAI.V35I5.16606},
  bibsource    = {dblp computer science bibliography, https://dblp.org}
}

@inproceedings{yang2025translationMEA,
  title={A Translation-Based Heterogeneous Graph Neural Network for Multiple Knowledge Graphs Alignment},
  author={Yang, Yaming and Luo, Zhuofeng and Wang, Zhe and Lu, Weigang and Lu, Yiheng and Guan, Ziyu and Zhao, Wei and Lv, Yuanhai},
  booktitle={Proceedings of ICDE},
  pages={2215--2226},
  year={2025}
}

@inproceedings{KoPA:MM24,
  author       = {Yichi Zhang and
                  Zhuo Chen and
                  Lingbing Guo and
                  Yajing Xu and
                  Wen Zhang and
                  Huajun Chen},
  title        = {Making Large Language Models Perform Better in Knowledge Graph Completion},
  booktitle    = {Proceedings of ACM {MM}},
  pages        = {233--242},
  year         = {2024},
  url          = {https://doi.org/10.1145/3664647.3681327},
  doi          = {10.1145/3664647.3681327},
  bibsource    = {dblp computer science bibliography, https://dblp.org}
}

@inproceedings{MKGL:NeurIPS24,
  author       = {Lingbing Guo and
                  Zhongpu Bo and
                  Zhuo Chen and
                  Yichi Zhang and
                  Jiaoyan Chen and
                  Yarong Lan and
                  Mengshu Sun and
                  Zhiqiang Zhang and
                  Yangyifei Luo and
                  Qian Li and
                  Qiang Zhang and
                  Wen Zhang and
                  Huajun Chen},
  title        = {{MKGL:} Mastery of a Three-Word Language},
  booktitle    = {Proceedings of NeurIPS},
  year         = {2024},
  url          = {http://papers.nips.cc/paper\_files/paper/2024/hash/fe03053bd2cf5b5c56de1e463bc53e1a-Abstract-Conference.html},
  bibsource    = {dblp computer science bibliography, https://dblp.org}
}

@INPROCEEDINGS{GLMKGC,
  author={He, Jinyan and Yang, Haitong},
  booktitle={Proceedings of ISCTIS}, 
  title={Multilingual Knowledge Graph Completion based on Global-Local Structure Encoding}, 
  year={2024},
  volume={},
  number={},
  pages={647-650},
  doi={10.1109/ISCTIS63324.2024.10699111}
}

@inproceedings{ICML23:multisource,
  author       = {Zequn Sun and
                  Jiacheng Huang and
                  Xiaozhou Xu and
                  Qijin Chen and
                  Weijun Ren and
                  Wei Hu},
  title        = {What Makes Entities Similar? {A} Similarity Flooding Perspective for
                  Multi-sourced Knowledge Graph Embeddings},
  booktitle    = {Proceedings of IMCL},
  pages        = {32875--32885},
  year         = {2023},
  url          = {https://proceedings.mlr.press/v202/sun23d.html},
  bibsource    = {dblp computer science bibliography, https://dblp.org}
}

@inproceedings{COLING25:DAEA,
  author       = {Linyan Yang and
                  Shiqiao Zhou and
                  Jingwei Cheng and
                  Fu Zhang and
                  Jizheng Wan and
                  Shuo Wang and
                  Mark Lee},
  title        = {{DAEA:} Enhancing Entity Alignment in Real-World Knowledge Graphs
                  Through Multi-Source Domain Adaptation},
  booktitle    = {Proceedings of COLING},
  pages        = {5890--5901},
  year         = {2025},
  url          = {https://aclanthology.org/2025.coling-main.393/},
  bibsource    = {dblp computer science bibliography, https://dblp.org}
}

@inproceedings{complEx,
  title={Complex embeddings for simple link prediction},
  author={Trouillon, Th{\'e}o and Welbl, Johannes and Riedel, Sebastian and Gaussier, {\'E}ric and Bouchard, Guillaume},
  booktitle={Proceedings of ICML},
  pages={2071--2080},
  year={2016},
  organization={PMLR}
}

@inproceedings{KEnS:chen-etal-2020-multilingual,
    title = "Multilingual Knowledge Graph Completion via Ensemble Knowledge Transfer",
    author = "Chen, Xuelu  and
      Chen, Muhao  and
      Fan, Changjun  and
      Uppunda, Ankith  and
      Sun, Yizhou  and
      Zaniolo, Carlo",
    booktitle = "Findings of EMNLP",
    month = nov,
    year = "2020",
    pages = "3227--3238",
}

@inproceedings{BootEA,
  author       = {Zequn Sun and
                  Wei Hu and
                  Qingheng Zhang and
                  Yuzhong Qu},
  title        = {Bootstrapping Entity Alignment with Knowledge Graph Embedding},
  booktitle    = {Proceedings of IJCAI},
  pages        = {4396--4402},
  year         = {2018},
  url          = {https://doi.org/10.24963/ijcai.2018/611},
  doi          = {10.24963/IJCAI.2018/611},
  bibsource    = {dblp computer science bibliography, https://dblp.org}
}

@inproceedings{DiffusionRM,
  author       = {Wenjie Wang and
                  Yiyan Xu and
                  Fuli Feng and
                  Xinyu Lin and
                  Xiangnan He and
                  Tat{-}Seng Chua},
  title        = {Diffusion Recommender Model},
  booktitle    = {Proceedings of SIGIR},
  pages        = {832--841},
  publisher    = {{ACM}},
  year         = {2023}
}

@inproceedings{DDPM:NIPS2020,
  author       = {Jonathan Ho and
                  Ajay Jain and
                  Pieter Abbeel},
  title        = {Denoising Diffusion Probabilistic Models},
  booktitle    = {Proceedings of NeurIPS},
  year         = {2020}
}

@inproceedings{DULNT,
  author       = {Jascha Sohl{-}Dickstein and
                  Eric A. Weiss and
                  Niru Maheswaranathan and
                  Surya Ganguli},
  title        = {Deep Unsupervised Learning using Nonequilibrium Thermodynamics},
  booktitle    = {Proceedings of ICML},
  volume       = {37},
  pages        = {2256--2265},
  year         = {2015},
}

@inproceedings{SDXL:ICLR2024,
  author       = {Dustin Podell and
                  Zion English and
                  Kyle Lacey and
                  Andreas Blattmann and
                  Tim Dockhorn and
                  Jonas M{\"{u}}ller and
                  Joe Penna and
                  Robin Rombach},
  title        = {{SDXL:} Improving Latent Diffusion Models for High-Resolution Image Synthesis},
  booktitle    = {Proceedings of ICLR},
  year         = {2024}
}

@inproceedings{LDM:CVPR2022,
  author       = {Robin Rombach and
                  Andreas Blattmann and
                  Dominik Lorenz and
                  Patrick Esser and
                  Bj{\"{o}}rn Ommer},
  title        = {High-Resolution Image Synthesis with Latent Diffusion Models},
  booktitle    = {Proceedings of CVPR},
  pages        = {10674--10685},
  year         = {2022},
  url          = {https://doi.org/10.1109/CVPR52688.2022.01042},
  doi          = {10.1109/CVPR52688.2022.01042},
  bibsource    = {dblp computer science bibliography, https://dblp.org}
}

@inproceedings{DreamRec:NeurIPS2023,
  author       = {Zhengyi Yang and
                  Jiancan Wu and
                  Zhicai Wang and
                  Xiang Wang and
                  Yancheng Yuan and
                  Xiangnan He},
  title        = {Generate What You Prefer: Reshaping Sequential Recommendation via Guided Diffusion},
  booktitle    = {Proceedings of NeurIPS},
  year         = {2023},
  url          = {http://papers.nips.cc/paper\_files/paper/2023/hash/4c5e2bcbf21bdf40d75fddad0bd43dc9-Abstract-Conference.html},
  bibsource    = {dblp computer science bibliography, https://dblp.org}
}

@inproceedings{DiffKG:WSDM24,
  author       = {Yangqin Jiang and
                  Yuhao Yang and
                  Lianghao Xia and
                  Chao Huang},
  title        = {DiffKG: Knowledge Graph Diffusion Model for Recommendation},
  booktitle    = {Proceedings of ACM WSDM},
  pages        = {313--321},
  publisher    = {{ACM}},
  year         = {2024},
  url          = {https://doi.org/10.1145/3616855.3635850},
  doi          = {10.1145/3616855.3635850},
  bibsource    = {dblp computer science bibliography, https://dblp.org}
}

@inproceedings{KGRS:WWW25,
  author       = {Huanyu Zhang and
                  Xiaoxuan Shen and
                  Baolin Yi and
                  Jianfang Liu and
                  Yinao Xie},
  title        = {A Plug-in Critiquing Approach for Knowledge Graph Recommendation Systems via Representative Sampling},
  booktitle    = {Proceedings of Web Conference},
  pages        = {322--333},
  publisher    = {{ACM}},
  year         = {2025},
  url          = {https://doi.org/10.1145/3696410.3714808},
  doi          = {10.1145/3696410.3714808},
  bibsource    = {dblp computer science bibliography, https://dblp.org}
}

@inproceedings{KAG:WWW25,
  author       = {Lei Liang and
                  Zhongpu Bo and
                  Zhengke Gui and
                  Zhongshu Zhu and
                  Ling Zhong and
                  Peilong Zhao and
                  Mengshu Sun and
                  Zhiqiang Zhang and
                  Jun Zhou and
                  Wenguang Chen and
                  Wen Zhang and
                  Huajun Chen},
  title        = {{KAG:} Boosting LLMs in Professional Domains via Knowledge Augmented
                  Generation},
  booktitle    = {Companion Proceedings of Web Conference},
  pages        = {334--343},
  publisher    = {{ACM}},
  year         = {2025},
  url          = {https://doi.org/10.1145/3701716.3715240},
  doi          = {10.1145/3701716.3715240},
  bibsource    = {dblp computer science bibliography, https://dblp.org}
}

@inproceedings{KGQA:WWW23,
  author       = {Lihui Liu and
                  Yuzhong Chen and
                  Mahashweta Das and
                  Hao Yang and
                  Hanghang Tong},
  title        = {Knowledge Graph Question Answering with Ambiguous Query},
  booktitle    = {Proceedings of Web Conference},
  pages        = {2477--2486},
  publisher    = {{ACM}},
  year         = {2023},
  url          = {https://doi.org/10.1145/3543507.3583316},
  doi          = {10.1145/3543507.3583316},
  bibsource    = {dblp computer science bibliography, https://dblp.org}
}

@inproceedings{FAAN:EMNLP20,
  author       = {Jiawei Sheng and
                  Shu Guo and
                  Zhenyu Chen and
                  Juwei Yue and
                  Lihong Wang and
                  Tingwen Liu and
                  Hongbo Xu},
  title        = {Adaptive Attentional Network for Few-Shot Knowledge Graph Completion},
  booktitle    = {Proceedings of EMNLP},
  pages        = {1681--1691},
  publisher    = {Association for Computational Linguistics},
  year         = {2020},
  url          = {https://doi.org/10.18653/v1/2020.emnlp-main.131},
  doi          = {10.18653/V1/2020.EMNLP-MAIN.131},
  bibsource    = {dblp computer science bibliography, https://dblp.org}
}

@article{FSKGC:TWEB24,
  author       = {Pengfei Luo and
                  Xi Zhu and
                  Tong Xu and
                  Yi Zheng and
                  Enhong Chen},
  title        = {Semantic Interaction Matching Network for Few-Shot Knowledge Graph
                  Completion},
  journal      = {{ACM} Trans. Web},
  volume       = {18},
  number       = {2},
  pages        = {20:1--20:19},
  year         = {2024},
  url          = {https://doi.org/10.1145/3589557},
  doi          = {10.1145/3589557},
  bibsource    = {dblp computer science bibliography, https://dblp.org}
}

@inproceedings{MultiVAE:WWW18,
  author       = {Dawen Liang and
                  Rahul G. Krishnan and
                  Matthew D. Hoffman and
                  Tony Jebara},
  title        = {Variational Autoencoders for Collaborative Filtering},
  booktitle    = {Proceedings of WWW},
  pages        = {689--698},
  publisher    = {{ACM}},
  year         = {2018},
  url          = {https://doi.org/10.1145/3178876.3186150},
  doi          = {10.1145/3178876.3186150},
  bibsource    = {dblp computer science bibliography, https://dblp.org}
}

@article{CFG:22,
  author       = {Jonathan Ho and
                  Tim Salimans},
  title        = {Classifier-Free Diffusion Guidance},
  journal      = {CoRR},
  volume       = {abs/2207.12598},
  year         = {2022},
  url          = {https://doi.org/10.48550/arXiv.2207.12598},
  doi          = {10.48550/ARXIV.2207.12598},
  eprinttype    = {arXiv},
  eprint       = {2207.12598},
  bibsource    = {dblp computer science bibliography, https://dblp.org}
}

@inproceedings{KGDM:AAAI24,
  author       = {Xiao Long and
                  Liansheng Zhuang and
                  Aodi Li and
                  Jiuchang Wei and
                  Houqiang Li and
                  Shafei Wang},
  title        = {{KGDM:} {A} Diffusion Model to Capture Multiple Relation Semantics for Knowledge Graph Embedding},
  booktitle    = {Proceedings of AAAI},
  pages        = {8850--8858},
  publisher    = {{AAAI} Press},
  year         = {2024},
  url          = {https://doi.org/10.1609/aaai.v38i8.28732},
  doi          = {10.1609/AAAI.V38I8.28732},
  bibsource    = {dblp computer science bibliography, https://dblp.org}
}

@article{DiffusionCom:25,
  author       = {Wei Huang and
                  Meiyu Liang and
                  Peining Li and
                  Xu Hou and
                  Yawen Li and
                  Junping Du and
                  Zhe Xue and
                  Zeli Guan},
  title        = {DiffusionCom: Structure-Aware Multimodal Diffusion Model for Multimodal
                  Knowledge Graph Completion},
  journal      = {CoRR},
  volume       = {abs/2504.06543},
  year         = {2025},
  url          = {https://doi.org/10.48550/arXiv.2504.06543},
  doi          = {10.48550/ARXIV.2504.06543},
  eprinttype    = {arXiv},
  eprint       = {2504.06543},
  bibsource    = {dblp computer science bibliography, https://dblp.org}
}

@article{MDKGP:24,
  author       = {Yichi Zhang and
                  Binbin Hu and
                  Zhuo Chen and
                  Lingbing Guo and
                  Ziqi Liu and
                  Zhiqiang Zhang and
                  Lei Liang and
                  Huajun Chen and
                  Wen Zhang},
  title        = {Multi-domain Knowledge Graph Collaborative Pre-training and Prompt
                  Tuning for Diverse Downstream Tasks},
  journal      = {CoRR},
  volume       = {abs/2405.13085},
  year         = {2024},
  url          = {https://doi.org/10.48550/arXiv.2405.13085},
  doi          = {10.48550/ARXIV.2405.13085},
  eprinttype    = {arXiv},
  eprint       = {2405.13085},
  bibsource    = {dblp computer science bibliography, https://dblp.org}
}

@article{DA-DIFF:TIP24,
  author       = {Duo Peng and
                  Qiuhong Ke and
                  Arulmurugan Ambikapathi and
                  Yasin Yazici and
                  Yinjie Lei and
                  Jun Liu},
  title        = {Unsupervised Domain Adaptation via Domain-Adaptive Diffusion},
  journal      = {{IEEE} Trans. Image Process.},
  volume       = {33},
  pages        = {4245--4260},
  year         = {2024},
  url          = {https://doi.org/10.1109/TIP.2024.3424985},
  doi          = {10.1109/TIP.2024.3424985},
  bibsource    = {dblp computer science bibliography, https://dblp.org}
}

@article{MMD:2012,
  author       = {Arthur Gretton and
                  Karsten M. Borgwardt and
                  Malte J. Rasch and
                  Bernhard Sch{\"{o}}lkopf and
                  Alexander J. Smola},
  title        = {A Kernel Two-Sample Test},
  journal      = {J. Mach. Learn. Res.},
  volume       = {13},
  pages        = {723--773},
  year         = {2012},
  url          = {https://dl.acm.org/doi/10.5555/2503308.2188410},
  doi          = {10.5555/2503308.2188410},
  bibsource    = {dblp computer science bibliography, https://dblp.org}
}

@inproceedings{SimCLR:2020,
  author       = {Ting Chen and
                  Simon Kornblith and
                  Mohammad Norouzi and
                  Geoffrey E. Hinton},
  title        = {A Simple Framework for Contrastive Learning of Visual Representations},
  booktitle    = {Proceedings of ICML},
  volume       = {119},
  pages        = {1597--1607},
  publisher    = {{PMLR}},
  year         = {2020},
  url          = {http://proceedings.mlr.press/v119/chen20j.html},
  bibsource    = {dblp computer science bibliography, https://dblp.org}
}

@inproceedings{KnowledgeExtrapolationSurvey:IJCAI23,
  author       = {Mingyang Chen and
                  Wen Zhang and
                  Yuxia Geng and
                  Zezhong Xu and
                  Jeff Z. Pan and
                  Huajun Chen},
  title        = {Generalizing to Unseen Elements: {A} Survey on Knowledge Extrapolation
                  for Knowledge Graphs},
  booktitle    = {Proceedings of IJCAI},
  pages        = {6574--6582},
  year         = {2023},
  url          = {https://doi.org/10.24963/ijcai.2023/737},
  doi          = {10.24963/IJCAI.2023/737},
  bibsource    = {dblp computer science bibliography, https://dblp.org}
}

@inproceedings{LDLang:NIPS23,
  author       = {Justin Lovelace and
                  Varsha Kishore and
                  Chao Wan and
                  Eliot Shekhtman and
                  Kilian Q. Weinberger},
  title        = {Latent Diffusion for Language Generation},
  booktitle    = {Proceedings of NeurIPS},
  year         = {2023},
  bibsource    = {dblp computer science bibliography, https://dblp.org}
}

@inproceedings{DiffLang:NIPS23,
  author       = {Ishaan Gulrajani and
                  Tatsunori B. Hashimoto},
  editor       = {Alice Oh and
                  Tristan Naumann and
                  Amir Globerson and
                  Kate Saenko and
                  Moritz Hardt and
                  Sergey Levine},
  title        = {Likelihood-Based Diffusion Language Models},
  booktitle    = {Proceedings of NeurIPS},
  year         = {2023},
  bibsource    = {dblp computer science bibliography, https://dblp.org}
}

@inproceedings{Seg:CVPR23,
  author       = {Aimon Rahman and
                  Jeya Maria Jose Valanarasu and
                  Ilker Hacihaliloglu and
                  Vishal M. Patel},
  title        = {Ambiguous Medical Image Segmentation Using Diffusion Models},
  booktitle    = {Proceedings of CVPR},
  pages        = {11536--11546},
  publisher    = {{IEEE}},
  year         = {2023},
  url          = {https://doi.org/10.1109/CVPR52729.2023.01110},
  doi          = {10.1109/CVPR52729.2023.01110},
  bibsource    = {dblp computer science bibliography, https://dblp.org}
}

@inproceedings{FactEmbedding:WWW24,
  author       = {Xiao Long and
                  Liansheng Zhuang and
                  Aodi Li and
                  Houqiang Li and
                  Shafei Wang},
  title        = {Fact Embedding through Diffusion Model for Knowledge Graph Completion},
  booktitle    = {Proceedings of WWW},
  pages        = {2020--2029},
  publisher    = {{ACM}},
  year         = {2024},
  url          = {https://doi.org/10.1145/3589334.3645451},
  doi          = {10.1145/3589334.3645451},
  bibsource    = {dblp computer science bibliography, https://dblp.org}
}

@inproceedings{DiffusionE:KDD24,
  author       = {Zongsheng Cao and
                  Jing Li and
                  Zigan Wang and
                  Jinliang Li},
  title        = {DiffusionE: Reasoning on Knowledge Graphs via Diffusion-based Graph
                  Neural Networks},
  booktitle    = {Proceedings of KDD},
  pages        = {222--230},
  publisher    = {{ACM}},
  year         = {2024},
  url          = {https://doi.org/10.1145/3637528.3671997},
  doi          = {10.1145/3637528.3671997},
  bibsource    = {dblp computer science bibliography, https://dblp.org}
}

@article{DiffDomainTea,
  author       = {Boyong He and
                  Yuxiang Ji and
                  Zhuoyue Tan and
                  Liaoni Wu},
  title        = {Diffusion Domain Teacher: Diffusion Guided Domain Adaptive Object
                  Detector},
  journal      = {CoRR},
  volume       = {abs/2506.04211},
  year         = {2025},
  url          = {https://doi.org/10.48550/arXiv.2506.04211},
  doi          = {10.48550/ARXIV.2506.04211},
  eprinttype    = {arXiv},
  eprint       = {2506.04211},
  bibsource    = {dblp computer science bibliography, https://dblp.org}
}

@inproceedings{DiffKD:NIPS23,
  author       = {Tao Huang and
                  Yuan Zhang and
                  Mingkai Zheng and
                  Shan You and
                  Fei Wang and
                  Chen Qian and
                  Chang Xu},
  title        = {Knowledge Diffusion for Distillation},
  booktitle    = {Proceedings of NeurIPS},
  year         = {2023},
  url          = {http://papers.nips.cc/paper\_files/paper/2023/hash/cdddf13f06182063c4dbde8cbd5a5c21-Abstract-Conference.html},
  bibsource    = {dblp computer science bibliography, https://dblp.org}
}

@inproceedings{sheng2026:IMKGC,
  author       = {Jiawei Sheng and Taoyu Su and Weiyi Yang  and Linghui Wang and Yongxiu Xu and  Tingwen Liu},
  title        = {Information-Theoretic Minimal Sufficient Representation for Multi-Domain Knowledge Graph Completion},
  booktitle    = {Proceedings of AAAI},
  year         = {2026}
}

\appendix

\section{Appendix}

In this appendix, we provide: (i) the dataset details, (ii) baselines, (iii) the evaluation protocol, and (iv) optimal hyper-parameters.

\subsection{Dataset Details}
For evaluation, we adopt three benchmarks with 14 KGs in our experiments: two multilingual datasets DBP-5L~\cite{Chen2017:MTransE}, E-PKG~\cite{Huang2022:SS-AGA}, and a constructed multi-domain dataset DWY~\cite{BootEA}.

\begin{itemize}[leftmargin=*]
    \item The \textbf{DBP-5L} dataset consists of 5 KGs extracted from DBpedia constructed in Greek (EL), English (EN), Spanish (ES), French (FR) and Japanese (JA).
    \item The \textbf{E-PKG} dataset is an e-commerce dataset about the mobile phone-related product information in 6 languages, including German (DE), English (EN), Spanish (ES), French (FR), Italian (IT) and Japanese (JA).
    \item The \textbf{DWY} dataset is a multi-domain dataset constructed in this paper based on~\citet{BootEA}, which includes DBpedia (DB), YAGO (YG) and Wiki (WK). We select 80\%, 10\%, 10\% triples of each KG as training, validation, testing data, respectively. To connect different KGs in the MKGC setting, we adopt the original aligned entities between paired KGs, and prepare equivalent entities between DB-YG, DB-WK and YG-WK. 
    We will release this dataset for future public research.
\end{itemize}
For all datasets, the equivalent entities are given to connect each of two KGs.
The relations are unified in a scheme across all KGs~\cite{Chen2017:MTransE, Tang2023:LSMGA}.
Detailed statistics of all datasets are shown in Table~\ref{Tab:data_sta}.

\begin{table}[!t]
\small
\centering
\caption{Statistics of DBP-5L~\cite{Chen2017:MTransE}, E-PKG~\cite{Huang2022:SS-AGA} and DWY~\cite{BootEA}.}
\setlength{\tabcolsep}{5pt}
\begin{tabular*}{0.48 \textwidth}{@{\extracolsep{\fill}}@{}l|crrrrr@{}}
\toprule
\bf Dataset & \bf KG & \bf \# Ent. & \bf \# Rel. & \bf \# Tra. & \bf \# Val. & \bf \# Tes. \\ 
\midrule
\multirow{5}{*}{\bf DBP-5L} & EL  & 5,231 & 111 & 8,670 & 4,152 & 1,017  \\
& EN  & 13,996 & 831 & 48,652 & 24,051 & 7,464 \\
& ES  & 12,382 & 144 & 33,036 & 16,220 & 4,810 \\
& FR  & 13,176 & 178 & 30,139 & 14,705 & 4,171 \\
& JA  & 11,805 & 128 & 17,979 & 8,633 & 2,162 \\
\midrule
\multirow{6}{*}{\bf E-PKG} & DE  & 17,223 & 21 & 45,515 & 22,753 & 7,602 \\
& EN  & 16,544 & 21 & 60,310 & 39,150 & 10,071 \\
& ES  & 9,595 & 21 & 18,090 & 9,039 & 3,034 \\
& FR  & 17,068 & 21 & 47,999 & 23,994 & 8022 \\
& IT  & 15,670 & 21 & 42,767 & 21,377 & 7,148 \\
& JA  & 2,642 & 21 & 10,013 & 5,002 & 1,688 \\
\midrule
\multirow{3}{*}{\bf DYW} & DB  & 23,315 & 180 & 85,506 & 10,688 & 10,780   \\
& YG  & 13,864 & 27 & 91,179 & 11,398 & 11,410 \\
& WK  & 17,743 & 90  & 82,047 & 10,255 & 10,296 \\
\bottomrule
\end{tabular*}
\label{Tab:data_sta}
\end{table}

\subsection{Baselines}
To evaluate our model, we select the following state-of-the-art methods as baselines:

(i) \textbf{Single-domain methods}, which learn and perform KGC independently on each KG:
\begin{itemize}[leftmargin=*]
\item  \textbf{TransE}~\cite{transE} models relations as translation in Euclidean space.
\item  \textbf{DisMult}~\cite{dismult} uses a bilinear function for semantic matching.
\item  \textbf{RotatE}~\cite{rotatE} represents relations as rotations in complex space.
\item  \textbf{KG-BERT}~\cite{KGbert} leverages pre-trained language models for KGC using textual descriptions of entities and relations.
\end{itemize}
These methods perform inference within individual KGs without knowledge transfer between KGs.

(ii) \textbf{Multi-domain methods}, which jointly leverage multiple KGs to enhance KGC through knowledge transfer:
\begin{itemize}[leftmargin=*]
\item  \textbf{KEnS}~\cite{Chen2020:KEnS} learns a unified embedding space across KGs and employs ensemble-based knowledge transfer.
\item  \textbf{CG-MuA}~\cite{CG_MuAlign} aligns KGs through a GNN with collective aggregation and adapts loss functions for multi-domain KGC.
\item  \textbf{AlignKGC}~\cite{Singh2021:AlignKGC} jointly performs KGC, entity alignment, and relation alignment across KGs.
\item  \textbf{SS-AGA}~\cite{Huang2022:SS-AGA} enhances multi-domain KGC by dynamically generating potential entity alignments.
\item  \textbf{LSMGA}~\cite{Tang2023:LSMGA} encodes KGs with an attentive relational graph encoder and fuses equivalent entities via attention mechanisms.
\item  \textbf{GLKGC}~\cite{GLMKGC} uses a transformer-based GNN for encoding, trained with KGC and entity alignment losses. 
\end{itemize}
These models mostly attempt consistency-based modules to align equivalent entities, and thus transfer knowledge from support KGs to benefit target KG predictions.

\subsection{Evaluation Protocol}
For generality, we evaluate the KGC model with the task of predicting tail entities.
For a query $(h, r, ?)$, we place all candidate tail entities in the query to form triples and measure the plausibility scores of the triples.
The tail entity in the triple with the highest score is treated as the final prediction of the tail entity.
Note that the entity set of a target KG serves as the corresponding candidate tail entities.
The testing data of all KGs are used to test the model, and we use the averaged metrics of all KGs to measure the overall performance~\cite{Tang2023:LSMGA}.
In detail, the following metrics are used:
\begin{itemize}[leftmargin=*]
    \item \textbf{Hits@N}: 
Hits@N (H@N for short) is the proportion of true entities that appear in the first $N$ entities of the sorted rank list. Hits@N can be defined as:
\begin{equation}
\begin{aligned}
\mathrm{Hits @ N}=\frac{1}{|\mathcal{Q}|} \sum_{q_i \in \mathcal{Q}} \mathbb{I}[\mathrm{rank}(i) \leq N],
\end{aligned}
\end{equation}
where $\mathcal{Q}$ denotes all query triples $(h, r, ?)$ in the testing data, $\mathrm{rank}_{i}$ denotes the rank position of the correct entity in the candidates for the $i$-th query, and $\mathbb{I}[\mathrm{rank}(i) \leq N]$ yields 1 if $i$ is ranked within top-$N$, and 0 otherwise.
This metric is bounded in the range [0, 1], where the higher, the better. 
Note that Hits@1 is equivalent to the precision in conventional classification tasks.
    \item \textbf{MRR}:
Mean reciprocal rank (MRR) measures the overall performance of the ranking, which is the average of the reciprocal ranks of results for all queries as:
\begin{equation}
\begin{aligned}
\mathrm{MRR}=\frac{1}{|\mathcal{Q}|} \sum_{q_i \in \mathcal{Q}} \frac{1}{\mathrm{rank}(i)},
\end{aligned}
\end{equation}
where $\mathcal{Q}$ also refers to the query triples $(h, r, ?)$. 
MRR is a useful metric since it reflects the overall ranks of all query triples. 
Higher MRR values indicate better performance, with 1 being the maximum achievable value.
\end{itemize}

\subsection{Optimal Hyper-parameters}
We implement our model with Pytorch\footnote{https://docs.pytorch.org/docs/stable/index.html} based on the PyG\footnote{https://pytorch-geometric.readthedocs.io/en/latest/} architecture.
Experiments are conducted on a server with Tesla T4 GPUs.
The optimal model is selected according to the average MRR of all KGs on their validation sets by grid-search with three trials~\cite{Tang2023:LSMGA}.
We use Adam~\cite{adam} to learn the model.
The tuning ranges of hyper-parameters are reported in the main text.
Here, we report the hyper-parameters used in Table~\ref{Tab:OptHypers} for re-implementation.

\begin{table}[!tbp]
\footnotesize
\centering\setlength{\tabcolsep}{5pt}
\begin{tabular*}{0.98\columnwidth}{@{\extracolsep{\fill}}@{}l|ccc@{}}
\toprule
\bf Hyper-parameters & \bf DBP-5L & \bf E-PKG & \bf DWY   \\ 
\midrule
batch size & 300 & 300 & 300  \\
learning rate & $1\times 10^{-3}$ & $1\times 10^{-3}$ & $1\times 10^{-3}$  \\
margin $\lambda$ & 0.5 & 0.5 & 0.5  \\
epoch & 30 & 50 & 30 \\
\midrule
embedding size $d$ & 256 & 256 & 256  \\
hidden dimension & 256 & 256 & 256  \\
encoder layer number $L$ & 2 & 2 & 2 \\
\midrule
ratio $p_u$ & 0.1 & 0.1 & 0.1 \\
strength $s$ & 2 & 1 & 2 \\
diffusion step $T$ & 16 & 32 & 16  \\
factor $\omega_1$ & $1\times 10^{-2}$ & $1\times 10^{-2}$ & $1\times 10^{-2}$ \\
factor $\omega_2$ & $1\times 10^{-3}$ & $1\times 10^{-3}$ & $1\times 10^{-3}$ \\
\bottomrule
\end{tabular*}
\caption{Detailed hyper-parameters of our DMKGC model.}
\label{Tab:OptHypers}
\end{table}

\subsection{Further Analysis (RQ3)}

\begin{figure}[!t]
\centering
\setlength{\tabcolsep}{0pt}
\begin{tabular}{ll}
 \includegraphics[width=0.5\columnwidth]{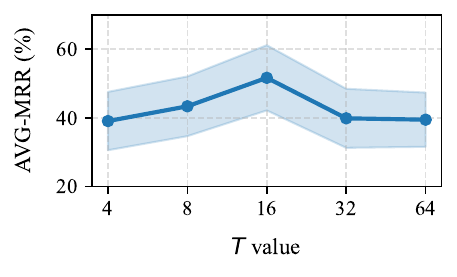}&
\includegraphics[width=0.5\columnwidth]{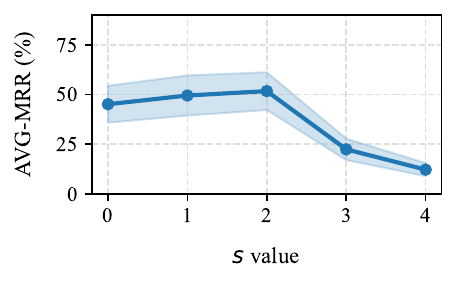}\\
\includegraphics[width=0.5\columnwidth]{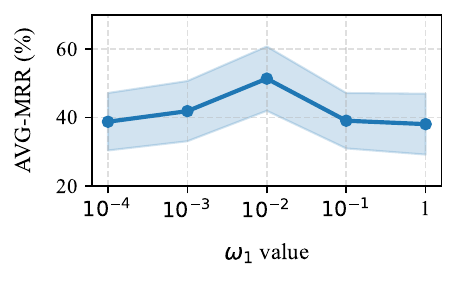} &
\includegraphics[width=0.5\columnwidth]{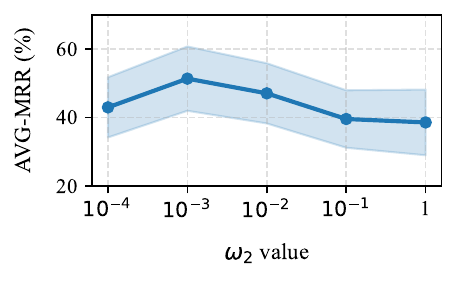} 
\end{tabular}
\caption{Impact of timestep number $T$, condition strength $s$, and harmonic factors $\omega_1, \omega_2$. Averaged MRR (\%) is reported with $\pm$ std of the 5 KGs on DBP-5L.}
\label{Exp:factors}
\end{figure}

\subsubsection{Impact on Hyper-parameters}
We show the impact of diffusion time step $T$, conditional strength factor $s$, and training balance factors $\omega_1$ and $\omega_2$, in Figure ~\ref{Exp:factors}.
We find that too large $T$ may not be helpful for relatively simplistic embedding generation.
A lower strength $s$ is better, in agreement with the expectation that the generation is unbiased towards the conditions.
Factors $\omega_1$ and $\omega_2$ require the taking of trade-off values to balance unbiased generation/regularization and the target task prediction.

\subsubsection{Comparison on Time Consuming}
To evaluate efficiency, we analyze the running time on DBP-5L.
In training, LSGMA, GLKGC and DMKGC (w/ 16 diffusion steps) spend 3.80, 4.87 and 4.71ms per triple, and achieve the best results at 49, 47, 16 training rounds, respectively.
This reflects that \textit{our diffusion-based model is still training-efficient, but can also accelerate convergence}.
In test time, LSGMA, GLKGC and DMKGC cost 5.80, 6.19, 6.64ms per triple, respectively. 
We believe \textit{the few reverse steps are controllable and acceptable considering remarkable accuracy improvements}.

\end{document}